\documentclass[11pt]{article}
\usepackage[letterpaper,margin=1in]{geometry}
\usepackage{times}
\usepackage{xspace}
\usepackage[dvipsnames]{xcolor}
\usepackage{graphicx,amsmath,amssymb,booktabs}
\usepackage[numbers,sort&compress]{natbib}
\usepackage[labelsep=period,font=small]{caption}
\usepackage[shortlabels,inline]{enumitem}
\usepackage{microtype}
\usepackage{multirow}
\usepackage{array}
\usepackage{colortbl}
\usepackage[section]{placeins}

\definecolor{lightgray}{gray}{0.93}

\makeatletter
\def\onedot{\ifx\@let@token.\else.\null\fi\xspace}
\makeatother
\def\eg{\emph{e.g}\onedot}  \def\etal{\emph{et al}\onedot}
\usepackage[breaklinks,colorlinks,allcolors=blue,pdftitle={What Does Hard-Negative Fine-Tuning Teach Multimodal Large Language Models? Controlled Evidence from a Compositional Benchmark},pdfauthor={Ranjith Chodavarapu}]{hyperref}
\title{What Does Hard-Negative Fine-Tuning Teach\\ Multimodal Large Language Models?\\[4pt] {\large Controlled Evidence from a Compositional Benchmark}}
\author{Ranjith Chodavarapu\\ Kent State University\\ \texttt{rchodava@kent.edu}}
\date{}
\begin{document}
\maketitle
\begin{abstract}
Fine-tuning on hard negatives is a popular way to improve the compositional reasoning of vision-language
models, but when accuracy goes up it is rarely clear what the model actually learned. We ask this question for
multimodal large language models. We use \textbf{CompHall}, 2,860 forced-choice questions (spatial, attribute,
counting and part--whole) with two graded hard negatives each, three model families (LLaVA-1.5, LLaVA-NeXT and
Idefics2), and controls that such studies often lack: gray-image evaluation, three seeds, option-order tests,
control arms trained on different question types, and independent benchmarks.
We find four things. First, CompHall is not a pure test of vision: part--whole questions can be answered from the
text alone (66\nobreakdash--95\% without the image), and per-image accuracy is unrelated to an object-hallucination probe
($r{\approx}{-}0.05$), so we see no evidence that compositional failures drive hallucination. Second, the original
fine-tuning method scored mostly image-placeholder tokens and hurt both LLaVA models. With a corrected loss,
fine-tuning raises accuracy on the spatial-relation questions of CV-Bench in all three families, by $+$3.2, $+$23.1
and $+$6.0 points (averaged over both option orders). Control arms show that the gain needs spatial training
questions, and for LLaVA-NeXT about two-thirds of it appears even on relation words that were only wrong options in
training. Third, the gain is not general: on CompHall itself it is significant only for LLaVA-NeXT ($+$6.9 points) and
not detectable for Idefics2, it reaches about 3 points at most on true/false statements, and LLaVA-1.5 fine-tuned on other
question types learns to answer ``true'' too often. Fourth, both losses raise the yes-bias of LLaVA-1.5 on POPE.
\end{abstract}

\section{Introduction}
\label{sec:intro}

Multimodal large language models (MLLMs), including LLaVA~\cite{liu2023llava}, InstructBLIP~\cite{dai2023instructblip}, and
GPT-4V~\cite{openai2023gpt4v}, can sometimes produce responses that do not accurately reflect the content of the input image.
This issue is commonly referred to as \emph{hallucination}~\cite{li2023evaluating,zhou2023analyzing,rohrbach2018object}. One
approach to addressing related limitations in vision-language models is fine-tuning with \emph{hard negatives}. These are
questions or captions with incorrect answers that differ from the correct ones in only one compositional detail, such as a
change in spatial relationships~\cite{yuksekgonul2023when,hsieh2023sugarcrepe}. Although this approach can improve
performance on compositional benchmarks, it is often unclear what the model actually learns from the process: improved visual
understanding, familiarity with the expected answer format, or patterns in how questions are phrased.

\paragraph{The question.}
We ask \textbf{what hard-negative fine-tuning teaches an MLLM}. To answer it, we built \textbf{CompHall}, a benchmark of
2,860 forced-choice questions in four compositional types (spatial, attribute, counting, part-whole), each with two
graded hard negatives (Section~\ref{sec:benchmark}), and we ran it on three model families with controls that such
studies often lack: gray-image (blind) evaluation, matched prompts and parsers, three training seeds, exact paired
tests, option-order tests, control arms that train on different question types, and independent benchmarks.

\paragraph{Motivation.}
Hard-negative fine-tuning is often motivated by the idea that compositional failures contribute to hallucination: the
inability to bind object identity, position, attributes, and part membership into one scene representation, as in the
\emph{binding problem} of perception research~\cite{treisman1980feature,treisman1996binding}. We tested this link
first. We found no evidence for it, so the rest of the paper asks the more modest question of what the fine-tuning does
teach.

\paragraph{Findings.}
\begin{enumerate}
\item \textbf{The benchmark is not a pure vision test, and compositional accuracy is not
      linked to hallucination} (Sections~\ref{sec:benchmark}--\ref{sec:analysis}). Six MLLMs,
      three of them recent, reach only 28\nobreakdash--51\% on spatial and 31\nobreakdash--45\% on counting
      questions. On 500 further part--whole questions all six models score 66\nobreakdash--95\% without
      the image, and under a second prompt every model is more accurate blind than with it.
      Per-image CompHall accuracy is uncorrelated with an object-hallucination probe
      (Pearson $r{\approx}{-}0.05$, 200 images).
\item \textbf{The original fine-tuning method had a flaw} (Section~\ref{sec:method}). In
      CCFT as first implemented, 97\% of the scored positions were image-placeholder tokens.
      That version degrades both LLaVA-1.5-7B and LLaVA-NeXT-7B on independent benchmarks
      (for example NeXT counting: 63.1\% to 12.6\%); a corrected loss scores answer tokens
      only.
\item \textbf{Spatial supervision transfers, but by very different amounts} (Sections~\ref{sec:generalization}--\ref{sec:vocab}).
      With the corrected loss, accuracy on the spatial-relation questions of the independent CV-Bench, averaged over both
      option orders, rises from 57.6\% to 60.8\% for LLaVA-1.5-7B, from 62.8\% to 85.9\% for LLaVA-NeXT-7B and from 63.8\%
      to 69.8\% for Idefics2-8B. Most of this needs spatial training questions: training on spatial questions alone gives
      64.5\%, 87.1\% and 68.7\%, and training on the other types alone 57.6\%, 66.1\% and 66.1\%. For the LLaVA models the
      gain holds on images from another dataset, and for LLaVA-NeXT about two-thirds of it appears even on relation words
      the model saw only as wrong options. The benchmark's own option order is misleading here: it makes LLaVA-1.5's gain
      look twice as large and Idefics2's look absent, because the models differ in which answer they prefer when unsure.
\item \textbf{The gain is not general.} It reaches counting only after spatial-only training (+4.2\,pp for LLaVA-NeXT-7B,
      significant in two of three seeds, and +3.9\,pp for Idefics2-8B, in all three; none for LLaVA-1.5-7B). On VSR's
      true/false statements it is about 3 points at most (LLaVA-NeXT-7B trained on spatial questions only; none for the
      other two models), and LLaVA-1.5-7B fine-tuned on other question types becomes biased toward answering ``true''
      ($-$3\,pp). On CompHall itself the gain differs across families: LLaVA-NeXT-7B improves by 6.9\,pp, LLaVA-1.5-7B
      does not improve significantly ($51.1\pm0.5$\% vs.\ 48.5\%) and loses the image's contribution to its spatial
      accuracy, and Idefics2-8B does not change (52.2\% to $52.6\pm0.4$\%). Both losses also raise the yes-bias of
      LLaVA-1.5-7B on POPE.
\end{enumerate}

\section{Related Work}
\label{sec:related}

\subsection{Multimodal Large Language Models}
MLLMs combine a pre-trained vision encoder (typically a CLIP ViT) with a
language model backbone via a lightweight connector module. LLaVA~\cite{liu2023llava}
uses a linear projection and LLaVA-1.5~\cite{liu2024improved} a two-layer MLP;
InstructBLIP~\cite{dai2023instructblip} and
BLIP-2~\cite{li2023blip2} use a Q-Former cross-attention bottleneck;
Idefics2~\cite{laurenccon2024idefics2} uses a perceiver resampler. LLaVA-NeXT
(also known as LLaVA-1.6)~\cite{liu2024llavanext} introduces dynamic
high-resolution visual encoding. Despite architectural diversity, all of these
models share the same fundamental limitation: the vision encoder produces a set
of independent patch tokens, and cross-modal binding is left to the LLM
attention heads without explicit structural inductive bias.

\subsection{Hallucination in MLLMs}
POPE~\cite{li2023evaluating} probes binary object existence and measures
yes-bias across three difficulty splits. MME~\cite{fu2023mme} evaluates a
broader set of visual capabilities. Rohrbach~\etal~\cite{rohrbach2018object}
measure caption hallucination in earlier vision-language models. AMBER~\cite{wang2023amber}
combines discriminative and generative hallucination evaluation.
Ramakrishnan~\etal~\cite{ramakrishnan2018overcoming} address language-prior shortcuts
in VQA. The hallucination benchmarks above measure object-existence or task-level accuracy and are not
designed around compositional binding.

\subsection{Compositional Visual Reasoning}
Visual compositional reasoning has been studied via CLEVR~\cite{johnson2017clevr},
GQA~\cite{hudson2019gqa}, and CREPE~\cite{ma2023crepe}. CREPE and
SugarCREPE~\cite{hsieh2023sugarcrepe} evaluate the compositionality of
image--text matching models with hard-negative captions, and
VALSE~\cite{parcalabescu2022valse} probes linguistic phenomena in
vision-language models. Fine-tuning on hard negatives of this kind was popularised for contrastive models by NegCLIP~\cite{yuksekgonul2023when}; we ask what it teaches an MLLM. CompHall instead asks multiple-choice questions about
scene-graph-derived compositional facts, gives each question two graded hard
negatives so that the option chosen on an error reveals the kind of mistake, and
relates question type to error rate.

\section{Background}
\label{sec:background}

\textbf{The MLLM pipeline and the binding bottleneck.} A vision encoder, typically a CLIP or
SigLIP ViT~\cite{radford2021clip}, maps the image to patch tokens; a connector projects them into
the language model's embedding space (a linear or MLP layer in LLaVA~\cite{liu2023llava,liu2024improved},
a Q-Former in BLIP-2 and InstructBLIP~\cite{li2023blip2,dai2023instructblip}, a perceiver resampler in
Idefics2~\cite{laurenccon2024idefics2});
and an LLM backbone generates text from the fused sequence. LLaVA-NeXT~\cite{liu2024llavanext} keeps the MLP connector
and adds multi-resolution tiling of the input image. In none of these architectures is a
component designed to bind features to objects: attributing a colour or a position to the right object
is left to self-attention layers trained on text. Perceptual psychology calls this the \emph{binding
problem}~\cite{treisman1980feature,treisman1996binding}: features processed separately must be
assigned to the right object, and failures appear as illusory conjunctions. We therefore expect
compositional errors to grow with scene complexity, although we do not test this directly
(Section~\ref{sec:clevr} measures only existence-question errors against object count). A schematic
and further detail are in Appendix~\ref{app:background}.

\textbf{Hallucination types.} We distinguish object-existence errors (T0; the target of POPE, which do not
require binding), attribute binding (T2), spatial relations (T1), part--whole and role binding (T4;
\eg, ``the pants are wearing the man''), and counting (T3, where the number of required bindings
grows with the number of instances). CompHall targets T1--T4 and excludes T0 by design: every question
concerns objects that are present in the image. A model can do well on a T0 benchmark (POPE accuracy
0.82\nobreakdash--0.87 for LLaVA-1.5-7B) while performing near chance on T1 and T3, as our results show.

\section{The CompHall Benchmark}
\label{sec:benchmark}

\subsection{Design Principles}
\label{sec:design}
CompHall is built around three design choices. \textbf{(i) Compositional focus.}
Every question requires relating at least two pieces of visual information (an
object and a position, colour, count, or second object); naming the queried
object alone is not sufficient. \textbf{(ii) Forced-choice format.} Each question
has three options: the correct answer and two hard negatives. Three-way forced
choice sets chance at 33.3\% and avoids the yes-bias of binary formats.
\textbf{(iii) Graded negatives.} The two negatives differ in how far they depart
from the correct answer. A \emph{near} negative changes one element (the opposite
direction, the colour of another annotated object, a count off by one, or a
swapped subject and object); a \emph{far} negative departs further (a different
spatial axis, an unrelated colour, a count off by two, or an unrelated object in
the scene). The option a model picks when it makes an error therefore shows what kind of
mistake it made.

\subsection{Question Types}
CompHall covers four question types corresponding to the types in
Section~\ref{sec:background}:
\begin{itemize}
\item \textbf{T1: Spatial relations.} \eg, \emph{Where is the person relative to
      the car?} Requires binding two objects to their positions; the options are
      spatial relations.
\item \textbf{T2: Attribute binding.} \eg, \emph{What color is the handle in the
      image?} Requires binding a colour to the queried object; the options are
      colours.
\item \textbf{T3: Object counting.} \eg, \emph{How many car can you see?}
      Requires identifying every instance of a category; the options are counts.
\item \textbf{T4: Part--whole relations.} \eg, \emph{Which statement correctly
      describes the relationship between the man and the pants?} Requires
      assigning the correct roles in a relation; the options are statements.
\end{itemize}

The four types differ in how many bindings a correct answer involves: T2 binds
one attribute to one object, T1 and T4 bind two objects to the roles of a
relation, and T3 requires identifying every instance of a category, so the number
of bindings grows with scene population. This grouping is a post-hoc reading of
Table~\ref{tab:main} (the heatmap is in Appendix~\ref{app:heatmap}), where T2 and T4 are answered
far more accurately than T1 and T3. It does not separate T1 from T4, which
involve similar numbers of bindings but differ widely in accuracy, so binding
count alone does not explain the results; part of the gap is that T4 can be
answered from the text alone (Section~\ref{sec:baseline}).

\subsection{Data Construction}
\textbf{Image sources.} Images are drawn from COCO val2017~\cite{lin2014coco}
(those with at least two annotated objects) and Visual Genome~\cite{krishna2017visual}
(the 2,000 images with the most annotated spatial and part--whole relationships, so
they are denser than typical scenes). The two sources map onto question
types: all T3 questions and most T1 questions (959 of 1,000 in the full file) use
COCO images, and all T2 and T4 questions use Visual Genome images, so question
type and image source are confounded (Table~\ref{tab:stats}). COCO val2017 is
held out from the standard COCO training split, but Visual Genome has no held-out
split, and we did not check overlap between the benchmark images and the training
data of the evaluated models.

\textbf{Question generation.} Questions are generated automatically from COCO and
Visual Genome annotations using templates. Each question receives two hard
negatives derived from the same annotations (Table~\ref{tab:negs}), and the
position of the correct option is randomised. An automatic filter removes
questions shorter than 15 characters, questions with duplicated options, and
questions containing giveaway words (\eg ``obviously''); each type is capped at
1,000 questions. No human verification was performed.

\textbf{Scale and splits.} The benchmark contains 2,860 questions in four types:
1,996 train, 441 validation, and 423 test (920, 199, and 200 images). Another 93
questions (70 train, 10 validation, 13 test, all with Visual Genome images) were removed because their image was not
found at the recorded path: our loader assigned some images to the wrong one of
Visual Genome's two image folders, although the files exist. No test
image appears in the training split. The training split is used for CCFT, the
validation split for checkpoint selection, and all reported accuracies are on the
test split.

\subsection{Baseline Evaluation}
\label{sec:baseline}
We evaluate three open-source MLLMs under zero-shot prompting
(Table~\ref{tab:main}). Responses are decoded greedily and matched to the closest
candidate by normalised edit distance.

\begin{table}[t]
\centering
\small
\caption{Zero-shot CompHall accuracy (\%) with the real image and with a uniform gray image (blind). Chance = 33.3\%.
$\dagger$: numerically below the 33.3\% chance level (not significant). $\ddagger$: second prompt wording (that of the CCFT evaluation). T4-L: 500 additional
part--whole questions on new images (95\% half-width at most 4.2\,pp).}
\label{tab:main}
\setlength{\tabcolsep}{2.6pt}
\begin{tabular}{lcccccc}
\toprule
Model & Overall & T1 & T2 & T3 & T4 & T4-L \\
\midrule
Chance                 & 33.3 & 33.3 & 33.3 & 33.3 & 33.3 & 33.3 \\
\midrule
LLaVA-1.5-7B           & 38.3 & 28.4$^\dagger$ & 56.2 & 30.9 & 58.1 & 67.0 \\
\quad blind            & 33.6 & 31.8 & 34.3 & 28.8 & 61.3 & 65.6 \\
LLaVA-NeXT-7B          & 52.5 & 42.6 & 79.0 & 38.1 & 74.2 & 79.8 \\
\quad blind            & 35.2 & 33.8 & 40.0 & 24.5 & 74.2 & 78.2 \\
Idefics2-8B            & 52.0 & 41.9 & 77.1 & 37.4 & 80.6 & 87.8 \\
\quad blind            & 41.6 & 39.9 & 54.3 & 25.9 & 77.4 & 90.0 \\
\midrule
LLaVA-1.5-7B$^\ddagger$ & 48.5 & 41.2 & 73.3 & 32.4 & 71.0 & 83.2 \\
\quad blind$^\ddagger$ & 39.2 & 32.4 & 50.5 & 30.2 & 74.2 & 90.2 \\
LLaVA-NeXT-7B$^\ddagger$ & 51.5 & 40.5 & 82.9 & 33.8 & 77.4 & 81.2 \\
\quad blind$^\ddagger$ & 39.0 & 39.2 & 49.5 & 21.6 & 80.6 & 82.6 \\
Idefics2-8B$^\ddagger$ & 54.6 & 43.9 & 74.3 & 43.2 & 90.3 & 89.8 \\
\quad blind$^\ddagger$ & 45.6 & 43.9 & 56.2 & 30.9 & 83.9 & 95.2 \\
\bottomrule
\end{tabular}
\end{table}

LLaVA-1.5-7B scores 28.4\% on T1 Spatial and 30.9\% on T3 Counting, numerically
below but statistically indistinguishable from the 33.3\% chance level
(two-sided binomial test, $p{=}0.22$ and $0.59$; $n{=}148$ and $139$).
LLaVA-NeXT-7B and Idefics2-8B score 42.6\% and 41.9\% on T1, modestly above
chance ($p{<}0.03$), and 38.1\% and 37.4\% on T3 ($p{>}0.2$), whereas their T2
and T4 accuracy is 74\nobreakdash--81\%.

\textbf{Blind baseline.} To see how much the questions depend on the image, we
repeat the evaluation with a uniform gray image (Table~\ref{tab:main}). Overall
accuracy falls from 38.3\% to 33.6\% for LLaVA-1.5-7B, from 52.5\% to 35.2\% for
LLaVA-NeXT-7B, and from 52.0\% to 41.6\% for Idefics2-8B. The image adds
22\nobreakdash--39\,pp on T2 and 2\nobreakdash--14\,pp on T3, but between $-$3 and $+$9\,pp on T1 and
between $-$3 and $+$3\,pp on T4. Under a second prompt wording (rows marked
$\ddagger$), overall accuracy changes by $+$10.2, about $-$1, and $+$2.6\,pp for
LLaVA-1.5-7B, LLaVA-NeXT-7B and Idefics2-8B, and the picture is the same: the image
adds 18\nobreakdash--33\,pp on T2, 2\nobreakdash--12\,pp on T3, 0\nobreakdash--9\,pp on T1, and $-$3 to $+$6\,pp on T4.
Across both prompts, part--whole questions are answered at 61\nobreakdash--84\% without the
image, and spatial questions gain little from it. The differences between types in
Table~\ref{tab:main} therefore partly reflect language priors rather than visual
binding.

\textbf{Recent models.} Three recent models (Qwen2.5-VL-7B, LLaVA-OneVision-7B,
InternVL3-8B) score 55\nobreakdash--59\% overall with the image under both prompts, but still
only 40\nobreakdash--51\% on spatial and 39\nobreakdash--45\% on counting questions (Table~\ref{tab:recent}). Without the image they score 37\nobreakdash--47\% overall and 64\nobreakdash--90\% on
part--whole questions, and the image adds 5\nobreakdash--10\,pp on spatial questions.

\begin{figure}[t]
\centering
\includegraphics[width=0.82\linewidth]{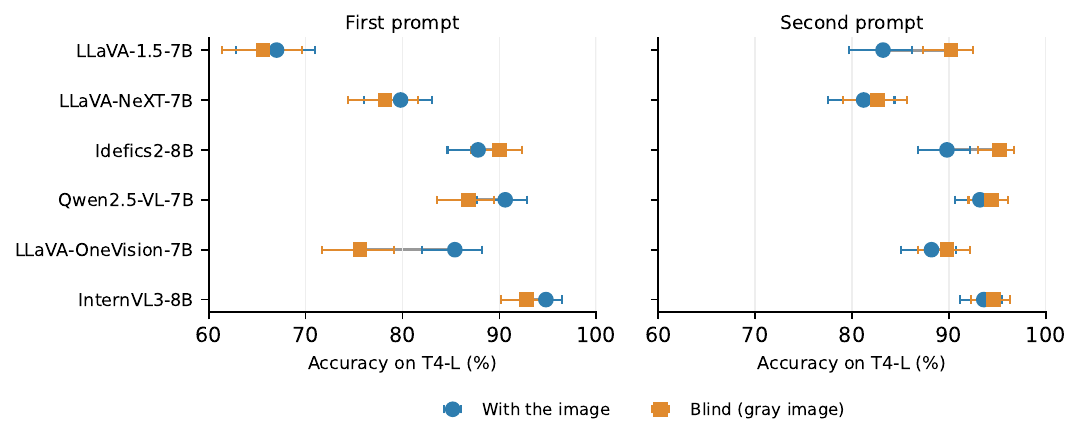}
\caption{Accuracy on the 500 part--whole questions of T4-L with the image and with a gray
image (blind), for six models under both prompts (bars: 95\% Wilson intervals). Under the
second prompt every model is at least as accurate blind as with the image.}
\label{fig:t4l}
\end{figure}

\textbf{A larger part--whole test set.} The part--whole test set has only 31
questions, so we built 500 further part--whole questions (T4-L) from Visual Genome
images that appear in no other split, with the same generator and filters and one
question per image (Appendix~\ref{app:stats}). All six models score 67\nobreakdash--95\% with
the image and 66\nobreakdash--95\% without it (Table~\ref{tab:main} and Table~\ref{tab:recent}). The image adds at most 9.8\,pp (LLaVA-OneVision, first prompt), and under
the second prompt every model is more accurate blind than with the image,
significantly so for LLaVA-1.5-7B (90.2\% vs.\ 83.2\%) and Idefics2-8B (95.2\% vs.\
89.8\%). These questions therefore measure language priors, not visual binding
(Figure~\ref{fig:t4l}).

\section{Compositional Errors and Hallucination}
\label{sec:analysis}

The benchmark results show a correlation between question type and error rate.
Question type is confounded with image source (T1 and T3 use COCO images, T2 and
T4 use Visual Genome images), so the type-level comparison below cannot separate
compositional difficulty from the effect of the image source. The synthetic-scene
study in Section~\ref{sec:clevr} varies the number of objects within a single
synthetic image source, but it probes object existence rather than binding.

\subsection{Type-Level Error Gap}
\label{sec:typegap}

\begin{figure}[t]
\centering
\includegraphics[width=0.82\linewidth]{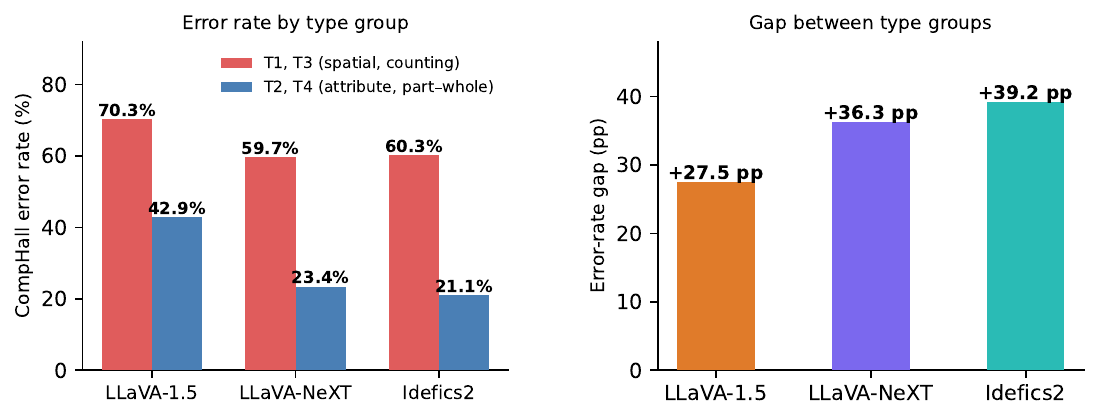}
\caption{Error-rate gap between hard (T1, T3) and easy (T2, T4) question types
per model; a group's error rate is the mean of its two types' error rates. The gap is 27.5\,pp for LLaVA-1.5-7B, 36.3\,pp for LLaVA-NeXT-7B,
and 39.2\,pp for Idefics2-8B.}
\label{fig:correlation}
\end{figure}

We partition CompHall questions into \emph{hard} types (T1 Spatial, T3 Counting,
requiring more simultaneous bindings) and \emph{easy} types (T2 Attribute, T4
Part) and compute the error-rate gap per model (Figure~\ref{fig:correlation}).

The gap widens as overall model accuracy increases (27.5\,pp, 36.3\,pp,
39.2\,pp for LLaVA-1.5-7B, LLaVA-NeXT-7B, and Idefics2-8B respectively),
indicating that stronger models improve on T2 and T4 while making smaller gains
on T1 and T3 --- consistent with compositional complexity having a structural
effect not reduced by general capability improvements. With only three models
the trend is descriptive rather than inferential, and the gap may partly reflect
image source rather than compositional difficulty.

\subsection{Per-Image Correlation with Object Hallucination}
\label{sec:corr}

We test the link more directly on 200 test images. For each image we compute
each model's accuracy on the CompHall questions about it, and a hallucination rate
from up to six yes/no existence questions asked of LLaVA-1.5-7B (``Is there a
\emph{object} in the image?''): three about objects named in the image's
annotations and three about objects that are not. The correlation is negligible
for all three models (Pearson $r{=}{-}0.05$, $p\geq0.44$; Spearman
$|\rho|\leq0.07$; Table~\ref{tab:corr}), with a 95\% interval for $r$ of about
$[-0.19, 0.09]$. The measurement is noisy: images have few questions, an object
missing from the annotations may still be in the image, and one probe model
serves all three rows, so only the first row compares a model with itself. The
result excludes moderate correlations but gives no support for a link between
per-image compositional accuracy and object hallucination.

\subsection{Synthetic-Scene Study}
\label{sec:clevr}

We draw synthetic scenes of 1\nobreakdash--5 flat coloured shapes in the style of
CLEVR~\cite{johnson2017clevr} (we do not use its renderer or images) and ask
LLaVA-1.5-7B two yes/no object-existence questions per scene, one about a present
and one about an absent colour (50 scenes per level; Appendix~\ref{app:synth}). The error rate rises from 6\% with one object to 42\% with five
(Figure~\ref{fig:causal}), and 200 independently generated 2-object and 3-object
scenes differ by 8.5\,pp ($t{=}3.39$, $p{=}0.0008$). These questions do not
require binding, so the study measures scene clutter rather than binding; it uses
one model and flat shapes.

\section{Compositional Contrastive Fine-Tuning}
\label{sec:method}

The benchmark results show that spatial and counting questions are hard and that
part of the errors are binding-consistent. We design a training objective
intended to reduce such errors. \textbf{Compositional Contrastive Fine-Tuning
(CCFT)} fine-tunes LLaVA-1.5-7B (Section~\ref{sec:generalization} adds two more models) on the CompHall training questions, using for each
question the correct answer and its two hard negatives (Section~\ref{sec:benchmark}).
It combines three terms.

\textbf{P1: Hard-negative contrast.} Standard supervised fine-tuning (SFT) trains
only on the correct answer and does not require the model to separate it from
plausible wrong ones. CCFT also requires the model to score the correct answer
higher than each hard negative by a margin.

\textbf{P2: Softmax ranking.} A cross-entropy over the scores of the correct answer
and the two hard negatives, with the correct answer as the target, moves
probability mass towards the correct answer and away from both negatives jointly.

\textbf{P3: Language-modelling term.} The standard language-modelling loss on the
correct-answer sequences, which is the SFT objective used as a baseline in
Section~\ref{sec:ablation}.

\subsection{Training Objective}

For a question $q$ about an image $x$ with correct answer $a^+$ and hard negatives
$a_1^-$ and $a_2^-$, we build the prompt ``\texttt{<image>} Question: $q$ Answer:
$a$'' for each candidate answer $a$. The score $s$ of an answer is the negative mean
token-level cross-entropy under the model. In our original implementation (v1) the
mean runs over the whole sequence and is computed for a mini-batch, so the terms
below compare mini-batch averages. In the corrected implementation (v2) it runs over
the answer tokens only and is computed per example. Let $s^+$ and $s_k^-$ denote the
scores of the correct answers and of the $k$-th hard negatives. The loss is

\begin{equation}
\mathcal{L}_\text{CCFT} = \mathcal{L}_\text{LM} + \lambda_c\,\mathcal{L}_\text{contrast} + \lambda_r\,\mathcal{L}_\text{rank}
\label{eq:ccft}
\end{equation}
with
\begin{align}
\mathcal{L}_\text{LM} &= -\,s^+ \label{eq:lm}\\
\mathcal{L}_\text{contrast} &= \tfrac{1}{2}\sum_{k=1}^{2}\max\!\left(0,\; m - s^+ + s_k^-\right) \label{eq:contrast}\\
\mathcal{L}_\text{rank} &= -\log\frac{e^{s^+}}{e^{s^+} + e^{s_1^-} + e^{s_2^-}} \label{eq:rank}
\end{align}
where $m{=}0.5$, $\lambda_c{=}0.5$, and $\lambda_r{=}0.3$.

\subsection{Implementation}

We apply CCFT to LLaVA-1.5-7B~\cite{liu2024improved} using LoRA~\cite{hu2022lora} on the attention
projections (rank 16, $\alpha{=}32$, dropout 0.05). Training runs for 3 epochs on
the CompHall training split (1,996 questions) with AdamW (learning rate
$2{\times}10^{-5}$, weight decay 0.01, cosine schedule with 10\% linear warmup),
batch size 2, fp16 autocast, and gradient clipping at 1.0. We keep the epoch with
the lowest validation loss, computed on 200 validation questions. All
hyperparameters are listed in Appendix~\ref{app:hyperparams}. We apply the corrected loss to LLaVA-NeXT-7B and
Idefics2-8B with the same LoRA rank, learning rate, schedule, epochs and loss weights and one question per
optimizer step (for Idefics2-8B, LoRA is on the language model only; Appendix~\ref{app:idefics}).
In v1 the score averages the token loss over all positions, 96.7\% of which are the
processor's image-placeholder tokens (40 training questions, plain model) and which
contribute 99.3\% of the loss; the correct-versus-negative score gap is $+$0.004 on
average and never reaches the margin of 0.5, so the contrastive and ranking terms
carry almost no information about the answers. v2 scores only the answer tokens, per
example (gap $+$0.83; margin already met for 47.5\% of pairs); everything else is
unchanged.

\section{What CCFT Teaches on CompHall}
\label{sec:experiments}

\subsection{CCFT vs.\ Baseline}
\label{sec:ccft_results}

Table~\ref{tab:ccft} compares LLaVA-1.5-7B with and without CCFT, using the same
prompt and parser for every run. The prompt matters: with the wording used here, the
plain model scores 48.5\% rather than the 38.3\% of Table~\ref{tab:main}, so about
10 of the 12.3\,pp between our first CCFT model and the Table~\ref{tab:main}
baseline come from the prompt alone. We report two implementations, each trained with
three seeds (42, 1, 2). The original (v1) averages the token loss over all positions,
96.7\% of which are image placeholders (Section~\ref{sec:method}); the corrected one
(v2) scores only the answer tokens, per example, with the same weights and margin.

With the image, v1 scores $49.2\pm1.9$\% and v2 $51.1\pm0.5$\%, against 48.5\% for
the plain model. No seed of either differs significantly from the plain model (exact
McNemar $p{\ge}0.47$ for v1; $p{=}0.09$, $0.25$, $0.15$ for v2, which fixes 29\nobreakdash--32
questions and breaks 19\nobreakdash--20 per seed). Without the image, v1 scores $44.8\pm1.5$\%
and v2 $44.6\pm0.5$\%, against 39.2\% ($p{\le}0.02$ for every v2 seed). The image
therefore adds 6.5\,pp after v2 fine-tuning (51.1\% vs.\ 44.6\%) against 9.3\,pp for
the plain model (48.5\% vs.\ 39.2\%): the fine-tuning improves accuracy without the
image more than with it. On CompHall we find no evidence that CCFT improves the visual binding of LLaVA-1.5-7B. By type, the plain model uses the image on spatial
questions (32.4\% blind vs.\ 41.2\% with the image), but after either fine-tuning
the image adds nothing (v2: 48.0\% vs.\ 47.3\%; Table~\ref{tab:pertype_next} and Figure~\ref{fig:imagegain}), so the
spatial gain of CCFT is a text effect. On the larger part--whole set (T4-L), v2 raises
LLaVA-1.5-7B from 83.2\% to $87.7\pm2.0$\% with the image (each seed $p{\le}0.0014$),
but the plain model is already more accurate blind (90.2\%) and v2 blind reaches
$91.1\pm0.8$\%, so here fine-tuning mostly removes the image's harm; v1 is unstable
($87.8\pm5.1$\%, seeds 82.8\nobreakdash--93.0\%; Table~\ref{tab:t4l}).

\begin{table}[t]
\centering
\small
\caption{CompHall accuracy (\%) of LLaVA-1.5-7B, with the image (overall and by type)
and without it (Blind). CCFT rows: mean over three seeds ($\pm$ standard deviation);
same prompt and parser throughout. Chance = 33.3\%.}
\label{tab:ccft}
\setlength{\tabcolsep}{3pt}
\begin{tabular}{lcccccc}
\toprule
Model & Overall & T1 & T2 & T3 & T4 & Blind \\
\midrule
Plain model & 48.5 & 41.2 & 73.3 & 32.4 & 71.0 & 39.2 \\
CCFT v1     & 49.2$\pm$1.9 & 41.9 & 65.7 & 36.7 & 83.9 & 44.8$\pm$1.5 \\
CCFT v2     & 51.1$\pm$0.5 & 47.3 & 72.4 & 32.6 & 79.6 & 44.6$\pm$0.5 \\
\bottomrule
\end{tabular}
\end{table}

\begin{figure}[t]
\centering
\includegraphics[width=0.82\linewidth]{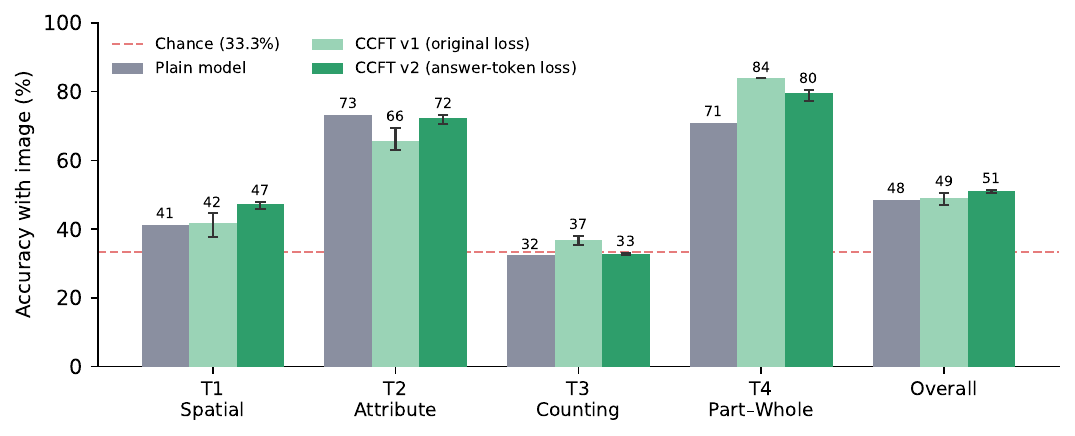}
\caption{Accuracy with the image for LLaVA-1.5-7B: plain model vs.\ CCFT v1 and v2
(means over three seeds; bars show the minimum--maximum range over seeds). The dashed line marks chance
(33.3\%).}
\label{fig:ccft}
\end{figure}

\subsection{Ablation Study}
\label{sec:ablation}

We ablate the loss terms with the corrected loss (seed 42; same budget, prompt and
parser as Table~\ref{tab:ccft}). SFT only (answer-token language modelling) scores
51.3\%, the contrastive term alone 51.8\%, the ranking term alone 52.7\%, and the full
objective 51.5\% (Table~\ref{tab:ablation}). All four lie within 1.4\,pp of each
other, and the seed-to-seed spread of the full objective is 0.5\,pp. Without the image
they score 45.2\nobreakdash--45.6\%. The ranking-only model is the only one that differs
significantly from the plain model with the image ($p{=}0.020$), which would not
survive correction for the seven comparisons we ran. The contrastive and ranking
terms therefore add nothing measurable to fine-tuning on the answers alone.

\begin{table}[t]
\centering
\small
\caption{Ablation of the loss terms (corrected loss, seed 42): overall accuracy (\%)
with and without the image, and the exact McNemar $p$ against the plain model with
the image.}
\label{tab:ablation}
\setlength{\tabcolsep}{4pt}
\begin{tabular}{lccc}
\toprule
Condition & Image & $p$ & Blind \\
\midrule
Plain model        & 48.5 & --    & 39.2 \\
SFT only           & 51.3 & 0.111 & 45.2 \\
SFT + Contrastive  & 51.8 & 0.076 & 45.4 \\
SFT + Ranking      & 52.7 & 0.020 & 45.6 \\
CCFT Full          & 51.5 & 0.092 & 45.2 \\
\bottomrule
\end{tabular}
\end{table}

\section{What Transfers: Independent Benchmarks and Controls}
\label{sec:generalization}

\subsection{CompHall Across Three Model Families}
\label{sec:arch}

\textbf{LLaVA-NeXT-7B.} We apply CCFT to it (the same prompt and parser as the baseline of this
run, which scores 53.0\% with the image and 39.0\% without it). The original loss,
one run, gives 52.5\% (fixes 69 questions, breaks 71; McNemar $p{=}0.93$) and 48.7\%
without the image: the image adds only 3.8\,pp instead of 14.0, and the model scores
100\% on T4 and 98.4\% on T4-L with the image (99.2\% blind). The corrected loss,
over three seeds, gives $59.8\pm0.4$\% ($+$6.9\,pp; $p{=}0.0056$, $0.0078$, $0.0015$
per seed) and $42.6\pm1.4$\% without the image ($+$3.6\,pp; $p{=}0.023$, $0.25$,
$0.23$). The gain is concentrated in spatial (44.6\% to 59.2\%) and counting (36.7\%
to 42.0\%) questions, where the image's contribution grows in every seed: the spatial
image gain is $+$9.5\,pp for the baseline against $+$14.2, $+$14.9 and $+$17.6\,pp, and
the counting gain $+$12.2\,pp against $+$18.2\,pp on average (Table~\ref{tab:pertype_next} in the
appendix; Figure~\ref{fig:imagegain}). This is unlike LLaVA-1.5-7B. On T4-L the corrected loss does
not help consistently ($82.1\pm2.7$\% vs.\ 80.6\%; $p{=}0.13$, $0.006$, $0.37$).

\textbf{Idefics2-8B.} To test a model with a different vision encoder, connector and language model
(SigLIP, perceiver resampler, Mistral-7B), we train Idefics2-8B with the corrected loss on all question
types, on the non-spatial types only and on spatial questions only, three seeds each (Appendix~\ref{app:idefics}).
Without fine-tuning it scores 52.2\% with the image and 45.9\% without it. Fine-tuning does not change
its CompHall accuracy: $52.6\pm0.4$\% (all types), $53.6\pm0.8$\% (non-spatial) and $51.2\pm0.5$\%
(spatial only), with McNemar $p\ge0.40$ in every seed, and even spatial-only training leaves its
spatial accuracy at $42.1\pm1.0$\% (plain: 42.6\%). Its blind accuracy rises slightly (to $48.8\pm1.0$\%
and $49.9\pm0.6$\% for the first two arms), mainly on counting (30.9\% to 39.3\% and 38.6\%; $p\le0.04$
in every seed) and attribute questions (52.4\% to 59.7\% and 61.9\%), which is a gain from text priors, not
from the image (Table~\ref{tab:idef_comp}). The in-distribution improvement is therefore clear in only one of the three families: it is significant for
LLaVA-NeXT-7B, small and not significant for LLaVA-1.5-7B, and not detectable for Idefics2-8B.

\begin{figure}[t]
\centering
\includegraphics[width=0.82\linewidth]{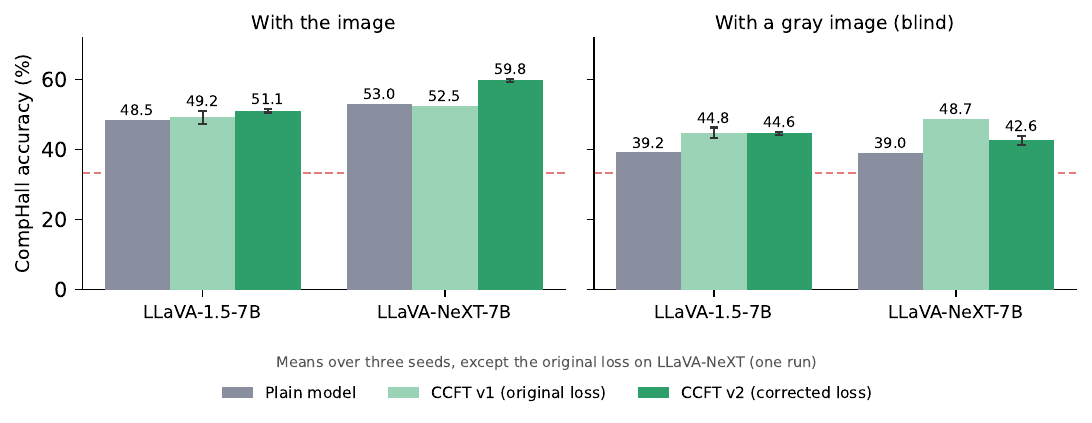}
\caption{Overall accuracy with the image (left) and with a gray image (right) for the plain
model and the two CCFT implementations. Means over three seeds (bars: standard
deviation), except the original loss on LLaVA-NeXT (one run).}
\label{fig:generalization}
\end{figure}

\begin{figure}[t]
\centering
\includegraphics[width=0.82\linewidth]{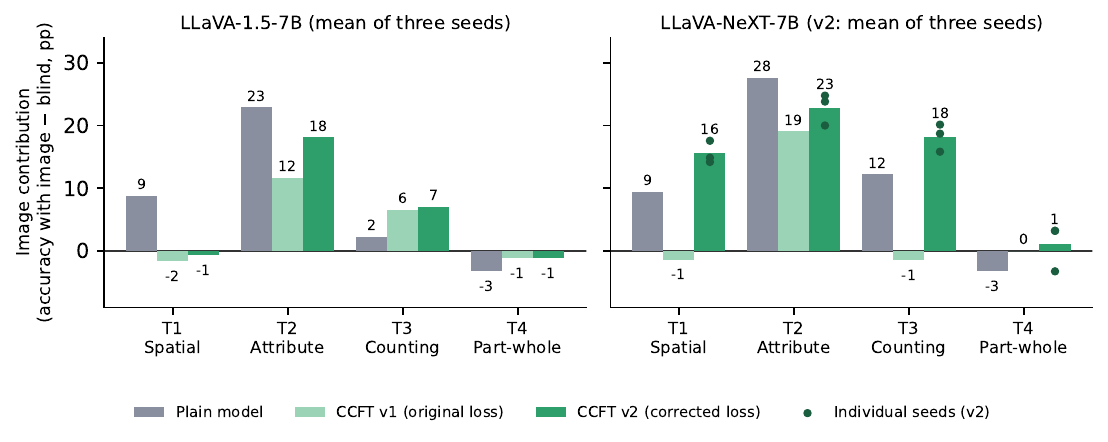}
\caption{Image contribution (accuracy with the image minus accuracy with a gray image) by
question type. In LLaVA-1.5-7B, fine-tuning removes the image's contribution to spatial
(T1) questions; in LLaVA-NeXT-7B, the corrected loss raises it for spatial and counting
(T3) questions. Bars are means over three seeds (LLaVA-NeXT original loss: one run); dots
are the individual LLaVA-NeXT v2 seeds.}
\label{fig:imagegain}
\end{figure}

\subsection{POPE Evaluation}
\label{sec:pope}

We evaluate on the POPE benchmark~\cite{li2023evaluating} ($n{=}444$ questions per
split) to measure how CCFT affects the general object-existence accuracy of
LLaVA-1.5-7B (Table~\ref{tab:pope}; the figure is in Appendix~\ref{app:popefig}); we did not run POPE for the
other models. POPE asks whether an object is present in the image. The probe is balanced, so a model with no bias answers ``yes'' about half of the time, and a yes-bias above 0.5 means that it claims absent objects too readily.

\begin{table}[t]
\centering
\small
\caption{POPE accuracy and yes-bias (fraction of ``yes'' answers) for the plain model,
CCFT v1 (original loss) and CCFT v2 (corrected loss), each the mean of three seeds.}
\label{tab:pope}
\setlength{\tabcolsep}{3pt}
\begin{tabular}{lcccccc}
\toprule
 & \multicolumn{3}{c}{Accuracy} & \multicolumn{3}{c}{Yes-bias} \\
\cmidrule(r){2-4}\cmidrule(r){5-7}
Split & Plain & v1 & v2 & Plain & v1 & v2 \\
\midrule
Random      & 0.867 & 0.725 & 0.893 & 0.381 & 0.748 & 0.494 \\
Popular     & 0.842 & 0.649 & 0.839 & 0.405 & 0.824 & 0.549 \\
Adversarial & 0.818 & 0.610 & 0.770 & 0.430 & 0.863 & 0.617 \\
\bottomrule
\end{tabular}
\end{table}

CCFT v1 reduces POPE accuracy on all three splits (adversarial: 0.818 to
$0.610\pm0.092$) and raises the yes-bias from 0.430 to $0.863\pm0.116$; it is also
unstable across seeds (accuracy standard deviation 0.09\nobreakdash--0.16). The corrected loss is
much less harmful and stable (standard deviation at most 0.009): accuracy changes by
$+$2.6\,pp (random), $-$0.3\,pp (popular) and $-$4.8\,pp (adversarial), but it still
shifts the model towards ``yes'' on every split (adversarial: 0.430 to
$0.617\pm0.021$; recall $+$13.9\,pp, precision $-$15.0\,pp). The shift does not come
from the contrastive or ranking terms: fine-tuning on the answers alone gives an
adversarial yes-bias of 0.588 (accuracy 0.773), against 0.586\nobreakdash--0.606 with either
added term and 0.617 with both (Table~\ref{tab:pope_full}).

\subsection{An Independent Benchmark: CV-Bench}
\label{sec:ext}
\label{sec:rev}

CompHall's questions share a generator with the fine-tuning data, so gains on it may reflect its
templates. We therefore evaluate the same models on the 2D tasks of CV-Bench~\cite{tong2024cambrian}:
counting and spatial-relation questions with 4\nobreakdash--6 and 2 options (chance 23\% and 50\%), on different
images and with different question writers. We drop the 160 questions whose COCO image appears in
CompHall, leaving 1,278 questions (723 counting and 555 spatial-relation questions), and use the prompts of the CompHall
evaluations, with and without the image (Appendices~\ref{app:ext} and~\ref{app:ctrl}). A model can gain or lose accuracy on a
two-option question simply by changing which letter or word it prefers, so we ask every
spatial-relation question twice, with the options in both orders, and report \emph{order-averaged}
accuracy (a model that always picks the same position scores exactly 50\%) and \emph{consistency}, the
share of questions answered with the same option content in both orders. Significance is a paired
bootstrap over questions for each seed. Counting has 4\nobreakdash--6 options and is reported in the benchmark's own order.

\begin{figure}[t]
\centering
\includegraphics[width=\linewidth]{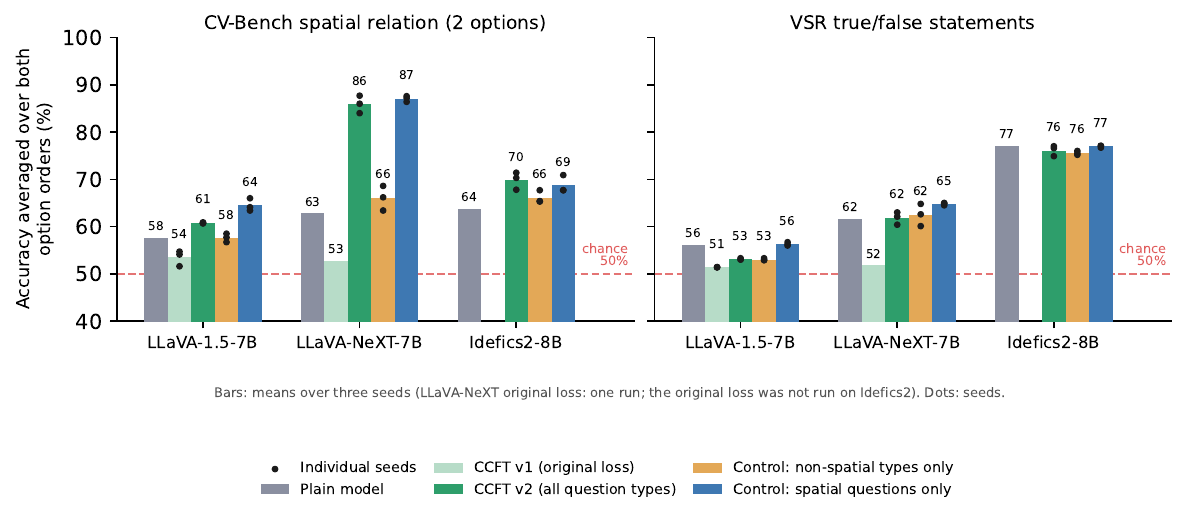}
\caption{Accuracy averaged over both option orders on CV-Bench spatial relations (left) and VSR (right), for
the plain models, both CCFT implementations (the original loss was run on the LLaVA models only) and the two
control arms. Bars are means over three seeds (LLaVA-NeXT original loss: one run), dots the individual seeds.}
\label{fig:rev}
\end{figure}

\textbf{Spatial relations.} The transfer is positive in all three families, but its size differs a lot
(Figure~\ref{fig:rev}; Table~\ref{tab:rev_rel}). Averaged over both option orders, accuracy with the image goes from
57.6\% to $60.8\pm0.1$\% for LLaVA-1.5-7B ($+$3.2 points), from 62.8\% to $85.9\pm1.9$\% for LLaVA-NeXT-7B ($+$23.1)
and from 63.8\% to $69.8\pm1.8$\% for Idefics2-8B ($+$6.0), and every seed is significant (paired bootstrap, $p\le0.002$).
LLaVA-NeXT's answers are also stable when the options are reordered (89\% consistent), and its gain is nearly the same in
the benchmark's own order ($+$24.6).

For the other two models the original order is misleading, in opposite directions. LLaVA-1.5-7B gains $+$6.7 points there
(60.0\% to $66.7\pm0.8$\%), about twice its order-averaged gain. Idefics2-8B gains only $+$0.8 (68.5\% to $69.3\pm2.4$\%,
not significant), because its plain model is nine points better in the original order than in the reversed one (68.5\% vs.\
59.1\%) and fine-tuning removes that asymmetry. The gain needs the image: blind accuracy stays between 48.8\% and 53.5\%
for the LLaVA models in every arm, and between 53.2\% and 58.7\% for Idefics2-8B (55.9\% for its plain model).

\textbf{The original loss degrades both LLaVA models.} Order-averaged, it lowers spatial-relation accuracy by
4.1\,pp (LLaVA-1.5-7B; $p\le0.049$ in every seed) and 10.0\,pp (LLaVA-NeXT-7B), and in the original order it lowers
counting from 53.3\% to $39.1\pm2.7$\% for LLaVA-1.5-7B ($p<0.0001$ in each seed) and from 63.1\% to 12.6\% for
LLaVA-NeXT-7B (chance 23\%; $p<0.0001$).

\textbf{Counting.} With the corrected loss on all types, counting does not improve in any family: 53.3\% to
$48.2\pm1.0$\% for LLaVA-1.5-7B ($p\le0.005$ in each seed), 63.1\% to $64.2\pm2.1$\% for LLaVA-NeXT-7B (not
significant) and 61.5\% to $59.0\pm0.8$\% for Idefics2-8B ($p=0.048$, $0.29$, $0.10$). Spatial-only training does
raise it for two of the three models (Section~\ref{sec:ctrl}).

\textbf{The blind control is informative only for spatial relations.} On a blank image the plain models answer
``0'' to 32\nobreakdash--99\% of the counting questions, and ``0'' is the correct answer for only 0.1\% of them, so their blind
counting accuracy (1\nobreakdash--22\%) lies below chance and says nothing about text priors. Fine-tuning changes this, which by
itself raises blind counting accuracy (LLaVA-1.5-7B: ``0'' in 16\nobreakdash--39\% of the answers of the original-loss models
against 70\% for the plain model; Idefics2-8B: blind counting accuracy rises from 7.7\% to 25\nobreakdash--31\% in the arms
trained on counting questions).

\textbf{The six plain models.} Accuracy with the image ranges from 49\% to 90\% on spatial relations and from 44\% to
73\% on counting (Table~\ref{tab:ext6}), and blind spatial accuracy stays at 49\nobreakdash--60\%, so, unlike CompHall's
part--whole questions, this benchmark cannot be answered without the image. The ranking of the six models agrees with
CompHall's for counting (Spearman $\rho=0.94$ and $0.60$ for the two prompts) and less for spatial questions
($\rho=0.37$ and $0.46$; $n{=}6$, so descriptive only).

\subsection{Where Does the Gain Come From?}
\label{sec:ctrl}

The corrected loss is trained on all four question types. To see which supervision the CV-Bench gain needs, we train
two control arms with the same loss, hyper-parameters and seeds: one on the non-spatial types only (T2--T4; 1,314
questions) and one on spatial questions only (T1; 682 questions), against 1,996 for the full set. All three models have
three seeds per arm (Appendices~\ref{app:ctrl} and~\ref{app:idefics}).

\textbf{Spatial supervision carries most of the gain.} Averaged over both orders, training on spatial questions alone raises
accuracy by 6.9, 24.3 and 4.9 points (LLaVA-1.5-7B, LLaVA-NeXT-7B, Idefics2-8B; significant in every seed), against 3.2,
23.1 and 6.0 points for all question types. The non-spatial types alone give 0.0, 3.3 and 2.3 points, that is 0\%, 14\% and
47\% of the spatial-only gain, and only three of the nine individual runs (three seeds for each of the three models) are significant (Table~\ref{tab:rev_rel}). Mixing the types
dilutes the transfer for LLaVA-1.5-7B but not for the other two models. In the benchmark's own order the same arms give
$+$1.9, $+$5.4 and $-$2.0 points (non-spatial) and $+$12.9, $+$24.9 and $+$2.1 points (spatial only; Table~\ref{tab:ctrl}).

\textbf{It holds on images from another dataset.} Split by image source, spatial-only training raises LLaVA-1.5-7B
from 61.9\% to 71.4\% on ADE20K images and from 58.0\% to 74.5\% on COCO images ($p\le0.001$ in every seed), and
LLaVA-NeXT-7B from 61.2\% to 86.6\% and from 67.4\% to 91.7\% (original order; Table~\ref{tab:src}); we did not split
Idefics2-8B's results by source.

\textbf{It reaches counting after spatial-only training in two of three families.} Spatial-only training raises counting
for LLaVA-NeXT-7B (63.1\% to $67.3\pm1.3$\%; $p=0.12$, $0.0002$, $0.005$) and for Idefics2-8B (61.5\% to
$65.4\pm0.4$\%; $p\le0.001$ in every seed), but for LLaVA-1.5-7B no arm does ($p>0.05$ for the spatial-only arm in every
seed). The other arms do not help counting in any family (LLaVA-NeXT-7B: all types $64.2\pm2.1$\%, non-spatial
$63.6\pm1.2$\%; Idefics2-8B: $59.0\pm0.8$\% and $58.6\pm0.7$\%).

\textbf{On CompHall the arms separate image-dependent from text-only gains for the LLaVA models.} Spatial-only training is
the only arm with a consistent image-conditioned spatial gain for LLaVA-1.5-7B ($+$11.1\,pp; $p\le0.03$ in each seed).
For LLaVA-NeXT-7B it raises spatial accuracy by $+$15.5\,pp with the image but only $+$3.6\,pp blind (three seeds), so the
image's contribution grows from $+$9.5 to $+$21.4\,pp, while the non-spatial arm leaves spatial accuracy unchanged
(44.6\%). All three LLaVA-1.5-7B arms raise its blind spatial accuracy ($p\le0.03$ in every seed;
Table~\ref{tab:pertype_next}). For Idefics2-8B no arm changes CompHall accuracy (Section~\ref{sec:arch}).

\textbf{Caveats.} The non-spatial arm's gain on LLaVA-NeXT-7B varies from $+$1.3 to $+$8.2\,pp across seeds in the
original order, which we cannot explain; the spatial-only arm of LLaVA-NeXT-7B also raises blind spatial-relation
accuracy slightly (52.3\% vs.\ 49.7\%; $p\le0.02$ in two of three seeds); and these controls cannot tell whether the
gain reflects spatial reasoning or the specific spatial vocabulary of CompHall's options (Section~\ref{sec:vocab}).

\subsection{True/False Statements and Answer Biases}
\label{sec:format}

On VSR~\cite{liu2023vsr}, true/false statements about spatial relations (1,218 questions; always answering ``true''
scores 51.4\%), the transfer is small at most. Averaged over both option orders, training on spatial questions alone gives
LLaVA-NeXT-7B a small but consistent gain (61.7\% to $64.8\pm0.3$\%, significant in every seed) and leaves LLaVA-1.5-7B
(56.1\% to 56.3\%) and Idefics2-8B (77.1\% to 77.0\%) unchanged. The other arms leave LLaVA-NeXT-7B unchanged, lower
Idefics2-8B slightly (by 1.0 and 1.5 points) and lower LLaVA-1.5-7B by about 3 points ($p<0.0001$ in every seed); the
original loss lowers both LLaVA models ($-$4.7 and $-$10.0 points). In the benchmark's own order the LLaVA drops look much
larger (6\nobreakdash--14 points) because of the answer biases described next (Tables~\ref{tab:vsr} and~\ref{tab:rev_vsr}).

\textbf{The biases differ by model.} In the original order option A is ``true'' and B is ``false'', and in the reversed order they swap; \emph{consistent} answers are those with the same meaning in both orders. LLaVA-1.5-7B follows the word: its plain model answers ``true'' for 74\% of the
statements in the original order and 98.5\% in the reversed order, and the fine-tuned models other than the spatial-only
ones for 92\nobreakdash--100\% in both, with 94\nobreakdash--99\% consistent answers, so fine-tuning on other question types has taught it an
affirmation bias. LLaVA-NeXT-7B's plain model answers ``A'' for 79\% and 85\% of the statements in the two orders (36\%
consistent answers), a position bias; only its original-loss model answers ``true'' regardless of order (consistency
99\%). Idefics2-8B's plain model is consistent (93\% of its answers stay the same when the order flips) with a mild
preference for ``true'' (62\% and 57\%).

\subsection{Held-Out Relation Words}
\label{sec:vocab}

We ran this test for the LLaVA models only. CompHall's spatial questions use a few direction words, and CV-Bench asks about the same words,
so the gain could reflect a learned mapping from those words to image layout rather than spatial
reasoning. We test this with two further arms (same loss, hyper-parameters and seeds; three seeds
each): one trained only on spatial questions whose correct answer is left or right (about 500
questions), the other only on those whose correct answer is above or below (158 questions). In each
arm the words of the other axis occur only as wrong options. We test on CV-Bench spatial relations
split by the axis of the correct answer (331 horizontal and 224 vertical questions, averaged over
both option orders), where both options of a question belong to the same axis, and on CompHall's own
spatial test questions (103 and 41; Appendix~\ref{app:vocab}).

\textbf{A large part of the gain transfers to words never seen as correct answers.} For
LLaVA-NeXT-7B the horizontal-only arm raises order-averaged accuracy on the held-out vertical
questions from 65.4\% to $78.8\pm0.8$\% ($+$13.4\,pp; $p<0.01$ in every seed), 66\% of the
spatial-only arm's gain on that axis (85.8\%), and the vertical-only arm raises the held-out
horizontal questions from 61.0\% to $78.2\pm2.7$\% ($+$17.2\,pp; $p<0.01$ in every seed), 64\% of
the spatial-only gain (87.9\%); with only 158 training questions it reaches almost the same
accuracy on the held-out axis as on its own (79.4\%; Table~\ref{tab:vocab_cv}). The picture for
LLaVA-1.5-7B is smaller and less consistent: the held-out gains are $+$2.6\,pp (vertical, 66.5\% to
69.1\%; significant in one of three seeds) and $+$3.9\,pp (horizontal, 51.5\% to 55.4\%;
significant in every seed), against $+$6.3 and $+$7.4\,pp for the spatial-only arm.

\textbf{But not all of it.} The trained axis does better than the held-out one for the
horizontal-only arm (85.9\% against 78.8\% for LLaVA-NeXT-7B), and in both LLaVA models the held-out
axis stays below the spatial-only arm. The gain is therefore partly general and partly specific to
the words seen as correct answers.

\textbf{CompHall's own test shows no transfer.} On CompHall no arm improves on the plain model on
its held-out axis (for example, LLaVA-NeXT-7B trained on left/right only scores 41.5\% against
46.3\% for the plain model on the 41 vertical questions; Table~\ref{tab:vocab_comp}). This is what
we would expect if the arms learn to reject words they only ever saw as wrong options: in CompHall
the held-out words appear in training only as wrong options, whereas on CV-Bench both options belong
to the held-out axis, so that shortcut cannot help. The vertical test set is small and we did not
test this explanation directly.

\section{Discussion}
\label{sec:discussion}

\paragraph{What hard-negative fine-tuning teaches.}
On CompHall itself, fine-tuning does not teach a general compositional skill. LLaVA-1.5-7B changes by $+$2.6 points with the image (not significant) and by $+$5.4 without it
($p{\le}0.02$ in every seed), and the image's contribution to its spatial questions disappears (from $+$8.8 to $-$0.7 points),
so what it gains comes more from the question text than from the image. LLaVA-NeXT-7B is
the exception: its spatial and counting accuracy rise and the image contributes more in all three seeds ($+$6.9 points
overall). Idefics2-8B does not change, apart from a rise in blind accuracy on counting and attribute questions that comes from
text priors. We therefore find no evidence that this kind of fine-tuning improves visual binding on CompHall for two of the
three models.

The independent benchmark tells a more positive but still limited story. Averaged over both option orders, the corrected loss
raises accuracy on CV-Bench's spatial-relation questions in all three families: strongly for LLaVA-NeXT-7B (62.8\% to
85.9\%), modestly for LLaVA-1.5-7B (57.6\% to 60.8\%) and Idefics2-8B (63.8\% to 69.8\%). The control arms show where the gain
comes from. Training on spatial questions alone gives $+$6.9, $+$24.3 and $+$4.9 points, the other question types alone give
$0.0$, $+$3.3 and $+$2.3 points, and all types together give $+$3.2, $+$23.1 and $+$6.0 points. For the LLaVA models the
spatial-only gain holds on images from ADE20K as well as COCO, and for LLaVA-NeXT-7B and Idefics2-8B it extends to counting
($+$4.2 and $+$3.9 points), but not for LLaVA-1.5-7B. Training on left/right questions only (or above/below only) gives
LLaVA-NeXT-7B about two-thirds of the spatial-only gain on the held-out axis ($+$13.4 and $+$17.2 points) and LLaVA-1.5-7B a
smaller, less consistent part ($+$2.6 and $+$3.9), but nothing on CompHall's own held-out questions, so the gain is partly
general and partly tied to the words seen as correct answers.

Two cautions apply. The gain is largely tied to the multiple-choice format: on VSR's true/false statements it is about 3
points at most and only for LLaVA-NeXT-7B, and models fine-tuned on other question types can lose up to 3 points
(LLaVA-1.5-7B becomes biased toward ``true''). And the benchmark's own option order can hide or exaggerate a gain, which is
why we report order-averaged accuracy.

The in-distribution and external results are compatible. Fine-tuning on CompHall's spatial questions can teach the question
format and its cues, which raise blind accuracy there, and, for LLaVA-NeXT-7B, also improves image-dependent spatial
judgements. The original loss is the opposite case: it teaches NeXT to answer part--whole questions from the text (100\% on
T4, 99.2\% blind on T4-L) and degrades both LLaVA models on the independent benchmark (NeXT counting: 63.1\% to 12.6\%).

\paragraph{Which wrong answer do models choose?}
Each CompHall question has a near and a far negative (Section~\ref{sec:benchmark}),
so the option chosen on an error shows what kind of mistake was made; if errors
were uniform over the two negatives, each would be chosen 50\% of the time
(counts in Table~\ref{tab:errdir}, Appendix~\ref{app:errdir}). On spatial questions (T1), LLaVA-NeXT-7B and
Idefics2-8B choose the opposite direction in 71\% (60/85) and 64\% (55/86) of their
errors under the first prompt ($p{<}0.001$ and $p{=}0.013$; two-sided binomial
test) and in 61\% and 64\% under the second ($p{=}0.042$ and $0.015$);
LLaVA-1.5-7B's errors are split evenly under the first prompt (51/106) and lean
towards the near negative under the second (53/87, $p{=}0.053$). On attribute
questions (T2), 62\nobreakdash--77\% of the first-prompt errors choose the colour of another
annotated object, which is significant only for LLaVA-NeXT-7B ($p{=}0.017$).
Counting errors split evenly between off-by-one and off-by-two (45\nobreakdash--52\%
off-by-one), and T4 has too few errors (5\nobreakdash--13 per model) to say anything. These
preferences are only partly image-dependent. Under the first prompt they vanish
with a gray image (45\nobreakdash--56\% near; all $p{>}0.28$), but under the second prompt
Idefics2-8B's preference persists without the image (T1: 66\% near vs.\ 64\% with
it, $p{=}0.004$; counting: 62.5\%, $p{=}0.018$), while it weakens for LLaVA-NeXT-7B
(52\% vs.\ 61\%) and LLaVA-1.5-7B (55\% vs.\ 61\%). A preference for the near
negative is therefore not specific evidence of visual binding errors; part of it is
a language prior. Models answer the same questions, so the tests are not
independent across models.

\paragraph{Scope of the hallucination claim.}
CompHall measures multiple-choice accuracy and the synthetic-scene study measures
yes/no existence errors; neither measures free-form hallucination, and the latter
does not involve binding. The one direct test of a link, the per-image correlation
(Section~\ref{sec:corr}), is null though noisy, and CCFT raises the POPE yes-bias. Our results show that compositional questions are hard and that some errors have the pattern binding failures would predict,
although part of that pattern is a language prior (previous paragraph); they do not support the claim that compositional
failures cause hallucination.

\paragraph{Implications for evaluating hard-negative fine-tuning.}
Four practices would have changed our conclusions, and we suggest them for similar studies. First, report accuracy without the
image: part--whole questions look like a vision test and are not. Second, for two-option tasks, ask every question in both
option orders: in the benchmark's own order the same fine-tuning looked twice as effective for one model and ineffective for
another. Third, train control arms on different question types, because a gain on a spatial benchmark may or may not need
spatial supervision. Fourth, check what the training loss actually scores: in our first implementation 97\% of the scored
positions were image-placeholder tokens, and the resulting models left CompHall's overall accuracy unchanged while degrading on
independent benchmarks.

\paragraph{Limitations.}
CompHall does not cover relational composition beyond the four types (transitive
relations, event structures) or temporal composition. Questions are template-generated and not human-verified. It is multiple-choice only, and question type is
confounded with image source (COCO for T1 and T3, Visual Genome for T2 and T4).
Visual Genome images are the 2,000 most densely annotated, not a random sample.
Several construction details limit what the types measure: T3 answers are COCO
annotation counts, which may miss small or occluded instances; the T2 near
negative is a random colour when an image has no other coloured object, and a
colour that appears elsewhere in the image may be chosen for that reason rather
than because of mis-binding; and the swapped statement in T4 is often implausible
as text, and a blind baseline confirms that language priors alone solve T4
(Section~\ref{sec:baseline}).
The full objective is trained with three seeds per implementation and each ablation
variant once, the original T4 test set has only 31 questions (T4-L adds 500), the original
loss was run once on LLaVA-NeXT-7B and not on Idefics2-8B, and all models have 7\nobreakdash--8B parameters and were fine-tuned with LoRA, so we cannot say
how the findings scale. Each seed is tested separately and without correction for multiple comparisons, and the paired tests treat
questions as independent, so the $p$-values are descriptive; three seeds also limit what we can say about seed-to-seed variance.
POPE was run for LLaVA-1.5-7B only, and CV-Bench's relation questions use only four direction words and two options. The external evaluation uses two benchmarks; blind
counting on CV-Bench is uninformative because models answer ``0'' for a blank image. The
controls leave a minority of the gain unexplained (0\nobreakdash--47\% of the spatial-only gain once option
order is controlled, depending on the model), the option-order control covers only the
two-option tasks, the held-out-vocabulary arms (LLaVA models only) are small (about 500 and 158 training questions) and CompHall's
held-out vertical test set has only 41 questions, and the gain extends to a true/false format only weakly. Results are prompt-sensitive: under a second prompt wording, overall accuracy moves from
38.3\% to 48.5\% for LLaVA-1.5-7B, from 52.5\% to 51.5\% for LLaVA-NeXT-7B, and from
52.0\% to 54.6\% for Idefics2-8B.

\section{Conclusion}
\label{sec:conclusion}

We asked what hard-negative fine-tuning teaches a multimodal large language model, and approached it with CompHall, blind
controls, matched prompts, three seeds, option-order tests, two independent benchmarks and three model families.

Spatial supervision transfers, but by very different amounts. With a corrected loss that scores only the answer tokens,
accuracy on CV-Bench's spatial-relation questions, averaged over both option orders, rises from 57.6\% to 60.8\% for
LLaVA-1.5-7B, from 62.8\% to 85.9\% for LLaVA-NeXT-7B and from 63.8\% to 69.8\% for Idefics2-8B. Most of the gain needs
spatial training questions (spatial only: 64.5\%, 87.1\% and 68.7\%; other types only: 57.6\%, 66.1\% and 66.1\%), for the
LLaVA models it holds on images from another dataset, and for LLaVA-NeXT-7B about two-thirds of it appears on relation words
that were only wrong options in training.

The gain is not general. On CompHall itself it is significant only for LLaVA-NeXT-7B ($+$6.9\,pp; LLaVA-1.5-7B: 51.1\% vs.\
48.5\%, not significant; Idefics2-8B: no change). It reaches counting only after spatial-only training, it is about 3 points at
most on true/false statements, and fine-tuning on other question types can make LLaVA-1.5-7B answer ``true'' too often.

The original method was flawed: its score was dominated by image-placeholder tokens, and it degraded both LLaVA models on
independent benchmarks. The contrastive and ranking terms add nothing over fine-tuning on the answers alone, and both losses
raise the POPE yes-bias, the original one unstably. The benchmark itself needs care: part--whole questions can be answered from
the text alone (66\nobreakdash--95\% without the image on 500 further questions), and per-image compositional accuracy is unrelated to an
object-hallucination probe ($r{\approx}{-}0.05$, noisy probe), so we find no evidence that compositional failures drive
hallucination. We did not test larger models, other training regimes or free-form hallucination.
CompHall and the CCFT code will be released.

\section*{Use of AI}
We used Claude (Anthropic), through its chat interface, throughout this project. It wrote first drafts and some revisions of the paper, including the abstract, background, related work, and discussion, as well as the \LaTeX{} source for those sections. It debugged and wrote some parts of the code, including the corrected-loss training scripts, the evaluation and statistical-analysis
scripts, and we used it to check the draft's numbers and logic against our result files. The author
designed the study, wrote much of the code, ran all experiments on local hardware, and every reported number comes from
the author's own result files. The author reviewed the full paper and takes responsibility for its content.

{\small
\bibliographystyle{ieeenat_fullname}
\bibliography{references}
}
\clearpage
\appendix
\clearpage

\section{CompHall Question Examples}
\label{app:examples}

Table~\ref{tab:examples} shows one test question per type with the correct answer
and the two hard negatives (near and far; Section~\ref{sec:design}).

\begin{table*}[h]
\centering
\small
\caption{CompHall test questions with the correct answer and the two hard negatives.
Text is quoted from the released data.}
\label{tab:examples}
\setlength{\tabcolsep}{4pt}
\begin{tabular}{p{1.6cm}p{5.0cm}p{2.3cm}p{2.5cm}p{2.7cm}}
\toprule
Type & Question & Correct & Near negative & Far negative \\
\midrule
T1 Spatial    & Where is the person relative to the car? & right of & left of & above \\
T2 Attribute  & What color is the handle in the image? & blue & green & orange \\
T3 Counting   & How many car can you see? & 11 & 10 & 13 \\
T4 Part--whole & Which statement correctly describes the relationship between the man and the pants? & The man is wearing the pants & The pants is wearing the man & The man is wearing the wii remote \\
\bottomrule
\end{tabular}
\end{table*}

\section{Per-Type Ablation Results}
\label{app:ablation}

Table~\ref{tab:ablation_pertype} extends the ablation of Table~\ref{tab:ablation} of the main paper with
a per-question-type breakdown (corrected loss, seed 42, with the image). The baseline
row is the plain model under the same prompt, so the $\Delta$ rows are matched
comparisons. All four trained conditions differ from each other by at most 1.4\,pp
overall, and the seed-to-seed spread of the full objective is 0.5\,pp (Table~\ref{tab:ccft} of the
main paper). Gains over the plain model are concentrated in T1 and T4; T4 has only 31
test questions, so $+$9.6\,pp is three questions.

An earlier ablation of the original loss, trained by a separate script (its settings
were not recorded; its defaults are 2 epochs on 500 questions) and evaluated with a
third prompt, scored 28.4\nobreakdash--46.3\% where the plain model scored 48.9\%. Because that
loss is dominated by image-placeholder tokens and its budget and prompt differ from
those of Table~\ref{tab:ccft}, we replaced it with the ablation above.

\begin{table}[h]
\centering
\small
\caption{Per-type accuracy (\%) with the image for the loss-term ablation (corrected
loss, seed 42), with delta relative to the plain model under the same prompt.}
\label{tab:ablation_pertype}
\setlength{\tabcolsep}{4pt}
\begin{tabular}{lrrrrr}
\toprule
Condition & T1 & T2 & T3 & T4 & Overall \\
\midrule
Plain model        & 41.2 & 73.3 & 32.4 & 71.0 & 48.5 \\
\midrule
SFT only           & 46.6 & 73.3 & 33.1 & 80.6 & 51.3 \\
\quad $\Delta$     & $+$5.4 & 0.0 & $+$0.7 & $+$9.6 & $+$2.8 \\
SFT + Contrastive  & 50.0 & 72.4 & 31.7 & 80.6 & 51.8 \\
\quad $\Delta$     & $+$8.8 & $-$0.9 & $-$0.7 & $+$9.6 & $+$3.3 \\
SFT + Ranking      & 50.7 & 73.3 & 33.1 & 80.6 & 52.7 \\
\quad $\Delta$     & $+$9.5 & 0.0 & $+$0.7 & $+$9.6 & $+$4.2 \\
CCFT Full          & 48.0 & 73.3 & 32.4 & 80.6 & 51.5 \\
\quad $\Delta$     & $+$6.8 & 0.0 & 0.0 & $+$9.6 & $+$3.0 \\
\bottomrule
\end{tabular}
\end{table}

\paragraph{With and without the image, both models.}
Table~\ref{tab:pertype_next} shows per-type accuracy with the image and with a gray image
for the baseline, both CCFT implementations (v1 = original loss, v2 = corrected loss) and the
two control arms of Section~\ref{sec:ctrl} (trained on the non-spatial types only or on spatial questions
only). LLaVA-1.5-7B rows are means over three seeds and use the prompt of Table~\ref{tab:ccft} of the main
paper; LLaVA-NeXT-7B rows use its own prompt, with v2 and both control arms means over three
seeds and the baseline and v1 single runs.

\begin{table}[h]
\centering
\small
\caption{Per-type accuracy (\%) with the image and blind (gray image).}
\label{tab:pertype_next}
\setlength{\tabcolsep}{4pt}
\begin{tabular}{llrrrrr}
\toprule
Input & Condition & T1 & T2 & T3 & T4 & Overall \\
\midrule
\multicolumn{7}{l}{\emph{LLaVA-1.5-7B (CCFT and controls: mean of three seeds)}} \\
Image & Baseline & 41.2 & 73.3 & 32.4 & 71.0 & 48.5 \\
      & CCFT v1  & 41.9 & 65.7 & 36.7 & 83.9 & 49.2 \\
      & CCFT v2  & 47.3 & 72.4 & 32.6 & 79.6 & 51.1 \\
      & Non-spatial only & 45.0 & 73.7 & 36.2 & 81.7 & 51.9 \\
      & Spatial only     & 52.3 & 69.5 & 32.1 & 68.8 & 51.1 \\
Blind & Baseline & 32.4 & 50.5 & 30.2 & 74.2 & 39.2 \\
      & CCFT v1  & 43.5 & 54.0 & 30.2 & 84.9 & 44.8 \\
      & CCFT v2  & 48.0 & 54.3 & 25.7 & 80.6 & 44.6 \\
      & Non-spatial only & 41.7 & 52.1 & 29.0 & 78.5 & 42.8 \\
      & Spatial only     & 46.4 & 46.7 & 27.8 & 68.8 & 42.0 \\
\midrule
\multicolumn{7}{l}{\emph{LLaVA-NeXT-7B (CCFT v2 and controls: mean of three seeds; baseline and v1: one run)}} \\
Image & Baseline & 44.6 & 79.0 & 36.7 & 77.4 & 53.0 \\
      & CCFT v1  & 41.2 & 78.1 & 34.5 & 100.0 & 52.5 \\
      & CCFT v2  & 59.2 & 79.0 & 42.0 & 77.4 & 59.8 \\
      & Non-spatial only & 44.6 & 79.7 & 42.7 & 89.2 & 55.9 \\
      & Spatial only     & 60.1 & 81.0 & 33.3 & 69.9 & 57.2 \\
Blind & Baseline & 35.1 & 51.4 & 24.5 & 80.6 & 39.0 \\
      & CCFT v1  & 42.6 & 59.0 & 36.0 & 100.0 & 48.7 \\
      & CCFT v2  & 43.7 & 56.2 & 23.7 & 76.3 & 42.6 \\
      & Non-spatial only & 39.9 & 58.1 & 23.5 & 81.7 & 42.1 \\
      & Spatial only     & 38.7 & 51.1 & 22.1 & 61.3 & 38.0 \\
\bottomrule
\end{tabular}
\end{table}

\section{Hyperparameter Details}
\label{app:hyperparams}

Table~\ref{tab:hyperparams} lists the CCFT hyperparameters used in all experiments,
as implemented in our training script.

\begin{table}[h]
\centering
\small
\caption{CCFT hyperparameters.}
\label{tab:hyperparams}
\begin{tabular}{ll}
\toprule
Parameter & Value \\
\midrule
Base model          & LLaVA-1.5-7B (\texttt{llava-hf}) \\
Fine-tuning method  & LoRA on q, k, v, o projections \\
LoRA rank / $\alpha$ / dropout & 16 / 32 / 0.05 \\
Optimiser           & AdamW, weight decay 0.01 \\
Learning rate       & $2\times10^{-5}$ \\
LR schedule         & Cosine, 10\% linear warmup \\
Epochs              & 3 \\
Batch size          & 2 (no gradient accumulation) \\
Optimisation steps  & 2,994 (998 per epoch) \\
Gradient clipping   & 1.0 \\
Precision           & fp16 autocast, fp32 LoRA weights \\
Margin $m$          & 0.5 \\
$\lambda_c$ / $\lambda_r$ & 0.5 / 0.3 \\
Score $s$           & v1: whole sequence, mini-batch; v2: answer tokens, per example \\
Checkpoint          & lowest validation loss (200 val.\ questions) \\
Random seed         & 42 \\
GPU                 & NVIDIA H100 NVL (95\,GB) \\
\bottomrule
\end{tabular}
\end{table}

\section{Extended POPE Analysis}
\label{app:pope}

Table~\ref{tab:pope_full} gives precision, recall and F1 for the POPE evaluation of
Section~\ref{sec:pope} ($n{=}444$ per split). v1 and v2 are means over three seeds. Both CCFT
implementations trade precision for recall; the corrected loss does so far less than
the original. The last three rows per split are single-seed variants of the corrected
loss (SFT only, contrastive term only, ranking term only); their accuracy and
yes-bias are 0.768\nobreakdash--0.779 and 0.586\nobreakdash--0.606 (adversarial), 0.833\nobreakdash--0.838 and 0.527\nobreakdash--0.536
(popular), and 0.885\nobreakdash--0.887 and 0.475\nobreakdash--0.486 (random), against 0.770 and 0.617,
0.839 and 0.549, and 0.893 and 0.494 for the full objective. Recall is identical across
splits for the plain model because the positive questions are shared.

\begin{table}[h]
\centering
\small
\caption{Full POPE results: precision, recall and F1.}
\label{tab:pope_full}
\setlength{\tabcolsep}{4pt}
\begin{tabular}{llccc}
\toprule
Split & Model & Precision & Recall & F1 \\
\midrule
Random      & Plain      & 0.982 & 0.748 & 0.849 \\
            & v1         & 0.672 & 0.973 & 0.789 \\
            & v2         & 0.898 & 0.887 & 0.893 \\
            & SFT only   & 0.905 & 0.860 & 0.882 \\
            & Contrastive & 0.906 & 0.865 & 0.885 \\
            & Ranking    & 0.898 & 0.874 & 0.886 \\
\midrule
Popular     & Plain      & 0.922 & 0.748 & 0.826 \\
            & v1         & 0.601 & 0.973 & 0.740 \\
            & v2         & 0.809 & 0.887 & 0.846 \\
            & SFT only   & 0.816 & 0.860 & 0.838 \\
            & Contrastive & 0.821 & 0.865 & 0.842 \\
            & Ranking    & 0.815 & 0.874 & 0.843 \\
\midrule
Adversarial & Plain      & 0.869 & 0.748 & 0.804 \\
            & v1         & 0.569 & 0.973 & 0.716 \\
            & v2         & 0.719 & 0.887 & 0.794 \\
            & SFT only   & 0.732 & 0.860 & 0.791 \\
            & Contrastive & 0.738 & 0.865 & 0.797 \\
            & Ranking    & 0.721 & 0.874 & 0.790 \\
\bottomrule
\end{tabular}
\end{table}

\noindent\textbf{Broader Impact.} CompHall is a diagnostic benchmark for
compositional errors in MLLMs, and CCFT is a fine-tuning study whose effect on
object hallucination is mixed (Section~\ref{sec:pope}). We do not anticipate direct harmful
applications.
Model checkpoints and benchmark data will be released.

\section{Error-Direction Counts}
\label{app:errdir}

Table~\ref{tab:errdir} gives, for each run, the number of incorrect answers that
choose the near or the far negative. Near/far: T1 opposite direction / different
axis; T2 colour of another object / unrelated colour; T3 off-by-one / off-by-two;
T4 subject--object swap / unrelated scene object. The middle block repeats the
evaluation with a gray image instead of the real one; the last two blocks use the second
prompt (the prompt of Table~\ref{tab:ccft} in the main paper): LLaVA-1.5-7B with and without CCFT,
and the other two baselines.

\begin{table}[h]
\centering
\small
\caption{Incorrect answers choosing the near / far negative.}
\label{tab:errdir}
\setlength{\tabcolsep}{3pt}
\begin{tabular}{lcccc}
\toprule
Run & T1 & T2 & T3 & T4 \\
\midrule
LLaVA-1.5-7B  & 51 / 55 & 29 / 17 & 49 / 47 & 6 / 7 \\
LLaVA-NeXT-7B & 60 / 25 & 17 / 5  & 39 / 47 & 3 / 5 \\
Idefics2-8B   & 55 / 31 & 15 / 9  & 45 / 42 & 1 / 5 \\
\midrule
LLaVA-1.5-7B, blind  & 47 / 54 & 35 / 34 & 51 / 48 & 7 / 5 \\
LLaVA-NeXT-7B, blind & 44 / 54 & 29 / 34 & 50 / 55 & 3 / 5 \\
Idefics2-8B, blind   & 50 / 39 & 27 / 21 & 60 / 43 & 2 / 5 \\
\midrule
Second prompt, LLaVA-1.5-7B: & & & & \\
plain, image & 53 / 34 & 21 / 7 & 51 / 43 & 4 / 5 \\
CCFT, image  & 50 / 34 & 25 / 7 & 38 / 50 & 1 / 4 \\
plain, blind & 55 / 45 & 30 / 22 & 52 / 45 & 5 / 3 \\
CCFT, blind  & 52 / 31 & 23 / 24 & 56 / 40 & 2 / 2 \\
\midrule
Second prompt, other baselines: & & & & \\
LLaVA-NeXT-7B, image & 54 / 34 & 12 / 6 & 42 / 50 & 3 / 4 \\
LLaVA-NeXT-7B, blind & 47 / 43 & 24 / 29 & 51 / 58 & 2 / 4 \\
Idefics2-8B, image   & 53 / 30 & 18 / 9 & 40 / 39 & 0 / 3 \\
Idefics2-8B, blind   & 55 / 28 & 27 / 19 & 60 / 36 & 1 / 4 \\
\bottomrule
\end{tabular}
\end{table}

\section{Benchmark Statistics}
\label{app:stats}

Table~\ref{tab:stats} gives the composition of the CompHall splits, which contain
the 2,860 questions used in all experiments. Questions are split by image: no test
image appears in the training split.

\begin{table}[h]
\centering
\small
\caption{Questions per split and type, with image source.}
\label{tab:stats}
\setlength{\tabcolsep}{4pt}
\begin{tabular}{llrrr}
\toprule
Type & Image source & Train & Val & Test \\
\midrule
T1 Spatial     & COCO (959 of 1,000)$^\ast$ & 682 & 164 & 148 \\
T2 Attribute   & Visual Genome & 471 & 92  & 105 \\
T3 Counting    & COCO          & 701 & 160 & 139 \\
T4 Part--whole & Visual Genome & 142 & 25  & 31  \\
\midrule
Total questions & & 1,996 & 441 & 423 \\
Images          & & 920   & 199 & 200 \\
\bottomrule
\end{tabular}

\vspace{2pt}
{\footnotesize $^\ast$ In the full file; the other T1 questions use Visual Genome images.}
\end{table}

Table~\ref{tab:negs} describes how the correct answer and the two hard negatives
are built for each type. Counts up to eight are written as words and larger counts
as digits, so the options of a counting question can mix the two forms. Option
positions are randomised.

\begin{table*}[h]
\centering
\small
\caption{Construction of the correct answer and the two hard negatives.}
\label{tab:negs}
\setlength{\tabcolsep}{4pt}
\begin{tabular}{p{1.9cm}p{4.0cm}p{4.2cm}p{4.2cm}}
\toprule
Type & Correct answer & Near negative & Far negative \\
\midrule
T1 Spatial & Annotated relation between two objects & The opposite relation (\eg left of $\to$ right of) & A random relation on another axis (\eg above) \\
T2 Attribute & Annotated colour of the queried object & Colour of another colour-annotated object in the image; a random colour if there is none & Another random colour \\
T3 Counting & Number of annotated instances of a category (at least two) & Count $\pm$1 & Count $\pm$2 (or $+$3 if it collides with the near negative) \\
T4 Part--whole & ``The S is P the O'' from an annotated part--whole or possession relation & Subject and object swapped & Object replaced by another object from the scene \\
\bottomrule
\end{tabular}
\end{table*}

\noindent\textbf{The larger part--whole set (T4-L).} We built 500 further part--whole
questions with the same generator, relationship tagging and quality filter as T4, with
two differences: they come from Visual Genome images that appear in no other split
(and are not COCO images used elsewhere), with at least three tagged relations and three
distinct objects, and there is one question per image. A question was also rejected if its
far negative was not an object of the scene. Of the 1,413 candidate images, the first
500 in a seeded random order that produced a question were used (18 images produced
none). The predicates are wearing (229), has (177), holding (56), sitting on (16),
attached to (8), standing on (6), mounted on (4), carrying (3) and part of (1); 437 of
the 500 subject--predicate--object triples are distinct, and the correct letter is
A, B and C for 176, 148 and 176 questions. T4-L is used for evaluation only.

\section{Per-Image Correlation}
\label{app:corr}

Table~\ref{tab:corr} gives the per-image correlation between CompHall accuracy and
the object-hallucination rate (Section~\ref{sec:corr}). Images are split into terciles of
compositional accuracy; the last three columns give the mean hallucination rate in
the lowest, middle and highest tercile. The hallucination probe is run with
LLaVA-1.5-7B for all three rows.

\begin{table}[h]
\centering
\small
\caption{Correlation between per-image CompHall accuracy and hallucination rate
(200 images). ANOVA compares the three terciles.}
\label{tab:corr}
\setlength{\tabcolsep}{3pt}
\begin{tabular}{lcccccc}
\toprule
Model & Pearson $r$ ($p$) & Spearman $\rho$ ($p$) & ANOVA $p$ & Low & Mid & High \\
\midrule
LLaVA-1.5-7B  & $-$0.053 (0.46) & $-$0.069 (0.33) & 0.40 & 0.112 & 0.080 & 0.085 \\
LLaVA-NeXT-7B & $-$0.055 (0.44) & $-$0.056 (0.43) & 0.75 & 0.106 & 0.094 & 0.085 \\
Idefics2-8B   & $-$0.052 (0.46) & $-$0.053 (0.45) & 0.053 & 0.116 & 0.061 & 0.116 \\
\bottomrule
\end{tabular}
\end{table}

\section{Synthetic-Scene Study}
\label{app:synth}

\begin{figure}[h]
\centering
\includegraphics[width=0.8\linewidth]{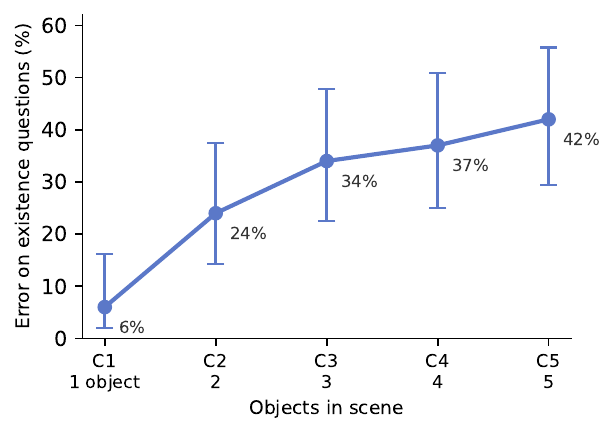}
\caption{Error rate of LLaVA-1.5-7B on object-existence questions in synthetic
scenes with 1\nobreakdash--5 objects (50 scenes per level, two questions each). Error bars are
95\% Wilson intervals over scenes.}
\label{fig:causal}
\end{figure}

Figure~\ref{fig:causal} shows the error rate of LLaVA-1.5-7B on the object-existence
questions of Section~\ref{sec:clevr}. Each level uses 50 scenes of flat coloured shapes and two
questions per scene (one about a colour that is present, one about a colour that is
absent), so the 100 questions per level come from 50 scenes; the intervals treat
the scene as the unit. The 95\% Wilson intervals are $[2.1, 16.2]$ at one object and
$[29.4, 55.8]$ at five, and those of two to five objects overlap. In a separate
comparison of 200 pairs of independently generated 2-object and 3-object scenes, the
mean difference in error rate is 8.5\,pp (one-sample $t$-test on the 200 differences,
$t{=}3.39$, $p{=}0.0008$).

\section{Recent Models}
\label{app:recent}

Table~\ref{tab:recent} gives CompHall accuracy for three recent open models, with the
real image and with a gray image (blind), under the two prompt wordings of the main
paper (Prompt~A: the wording of Table~\ref{tab:main}; Prompt~B: the wording of Table~\ref{tab:ccft}). Second-prompt
responses are scored with a parser that also accepts answers of the form ``(B) right of''.

\begin{table}[h]
\centering
\small
\caption{CompHall accuracy (\%) of three recent models. Chance = 33.3\%. T4-L: 500 additional part--whole questions.}
\label{tab:recent}
\setlength{\tabcolsep}{3pt}
\begin{tabular}{llccccccc}
\toprule
Model & Prompt & Input & Overall & T1 & T2 & T3 & T4 & T4-L \\
\midrule
Qwen2.5-VL-7B   & A & image & 58.6 & 50.7 & 84.8 & 41.0 & 87.1 & 90.6 \\
                & A & blind & 46.8 & 43.9 & 57.1 & 33.8 & 83.9 & 86.8 \\
                & B & image & 58.9 & 50.0 & 89.5 & 38.8 & 87.1 & 93.2 \\
                & B & blind & 46.1 & 44.6 & 59.0 & 30.2 & 80.6 & 94.4 \\
\midrule
LLaVA-OneVision-7B & A & image & 55.6 & 47.3 & 82.9 & 39.6 & 74.2 & 85.4 \\
                & A & blind & 37.1 & 37.2 & 41.0 & 28.1 & 64.5 & 75.6 \\
                & B & image & 56.0 & 46.6 & 81.9 & 41.7 & 77.4 & 88.2 \\
                & B & blind & 42.8 & 37.8 & 55.2 & 30.9 & 77.4 & 89.8 \\
\midrule
InternVL3-8B    & A & image & 56.5 & 40.5 & 83.8 & 45.3 & 90.3 & 94.8 \\
                & A & blind & 39.2 & 33.8 & 52.4 & 23.7 & 90.3 & 92.8 \\
                & B & image & 55.3 & 41.2 & 85.7 & 39.6 & 90.3 & 93.6 \\
                & B & blind & 40.7 & 35.8 & 58.1 & 21.6 & 90.3 & 94.6 \\
\bottomrule
\end{tabular}
\end{table}

\section{CCFT on the Larger Part--Whole Set}
\label{app:t4l}

Table~\ref{tab:t4l} gives accuracy on the 500 questions of T4-L with the image and with a
gray image for the plain models and each CCFT seed, with the exact McNemar $p$-value
against the plain model on the same questions. LLaVA-1.5-7B uses the prompt of Table~\ref{tab:ccft}
of the main paper; LLaVA-NeXT-7B uses its own prompt.

\begin{table}[h]
\centering
\small
\caption{Accuracy (\%) on T4-L ($n{=}500$) and McNemar $p$ against the plain model.}
\label{tab:t4l}
\setlength{\tabcolsep}{4pt}
\begin{tabular}{llcrcr}
\toprule
Model & Condition & Image & $p$ & Blind & $p$ \\
\midrule
LLaVA-1.5-7B & Plain      & 83.2 & --    & 90.2 & --    \\
             & v1, seed 42 & 87.6 & 0.004 & 90.2 & 1.00 \\
             & v1, seed 1  & 93.0 & $<$0.0001 & 91.6 & 0.43 \\
             & v1, seed 2  & 82.8 & 0.91  & 84.2 & 0.001 \\
             & v2, seed 42 & 90.0 & $<$0.0001 & 92.0 & 0.09 \\
             & v2, seed 1  & 86.4 & 0.0009 & 91.0 & 0.39 \\
             & v2, seed 2  & 86.8 & 0.0014 & 90.4 & 1.00 \\
\midrule
LLaVA-NeXT-7B & Plain      & 80.6 & --    & 83.8 & --    \\
              & v1         & 98.4 & $<$0.0001 & 99.2 & $<$0.0001 \\
              & v2, seed 42 & 83.0 & 0.13  & 85.8 & 0.17 \\
              & v2, seed 1  & 84.2 & 0.006 & 86.8 & 0.02 \\
              & v2, seed 2  & 79.0 & 0.37  & 83.2 & 0.78 \\
\bottomrule
\end{tabular}
\end{table}

\section{The External Benchmark (CV-Bench)}
\label{app:ext}

We use the 2D tasks of CV-Bench (788 counting and 650 spatial-relation questions; the
counting questions have 4\nobreakdash--6 options and always include ``0''), without the 160 questions
whose COCO image appears in CompHall (n = 1,278; chance 23\% for counting and 50\% for
spatial relations). Prompts follow the CompHall evaluations, with only the list of letters
adapted to each question; responses are scored with a parser restricted to the valid
letters. Table~\ref{tab:ext} gives the fine-tuning results, with the exact McNemar tests of
the main text computed per task and seed against the plain model; Table~\ref{tab:ext6} gives
the six plain models. A blank image leads the models to answer ``0'' (the correct answer
for 0.1\% of the questions), so blind counting accuracy is not a measure of text priors.

\begin{table}[h]
\centering
\small
\caption{CV-Bench accuracy (\%): overall with the image, and by task with and without the
image. v1 and v2 are means over three seeds, except LLaVA-NeXT v1 (one run).}
\label{tab:ext}
\setlength{\tabcolsep}{3.5pt}
\begin{tabular}{llccccc}
\toprule
 & & Overall & \multicolumn{2}{c}{Image} & \multicolumn{2}{c}{Blind} \\
\cmidrule(r){4-5}\cmidrule(r){6-7}
Model & Condition & image & Count & Relation & Count & Relation \\
\midrule
LLaVA-1.5-7B  & Plain & 56.2 & 53.3 & 60.0 & 16.0 & 49.7 \\
              & CCFT v1 & 47.0 & 39.1 & 57.3 & 36.1 & 53.5 \\
              & CCFT v2 & 56.2 & 48.2 & 66.7 & 23.5 & 49.7 \\
\midrule
LLaVA-NeXT-7B & Plain & 63.5 & 63.1 & 64.1 & 4.4 & 49.7 \\
              & CCFT v1 & 28.9 & 12.6 & 50.1 & 19.4 & 48.8 \\
              & CCFT v2 & 74.9 & 64.2 & 88.7 & 5.2 & 51.5 \\
\bottomrule
\end{tabular}
\end{table}

\begin{table}[h]
\centering
\small
\caption{The six plain models on CV-Bench (\%), under both prompts. ``0'': share of blank-image
counting questions answered ``0''.}
\label{tab:ext6}
\setlength{\tabcolsep}{3.5pt}
\begin{tabular}{llccccc}
\toprule
 & & \multicolumn{2}{c}{Image} & \multicolumn{2}{c}{Blind} & \\
\cmidrule(r){3-4}\cmidrule(r){5-6}
Model & Prompt & Count & Relation & Count & Relation & ``0'' \\
\midrule
LLaVA-1.5-7B       & A & 44.3 & 49.0 & 21.9 & 49.0 & 31.5 \\
                   & B & 53.3 & 60.0 & 16.0 & 49.7 & 69.8 \\
LLaVA-NeXT-7B      & A & 62.5 & 57.7 & 10.2 & 50.3 & 80.6 \\
                   & B & 60.2 & 62.2 & 4.8  & 50.5 & 90.6 \\
Idefics2-8B        & A & 62.4 & 64.7 & 4.4  & 58.7 & 91.6 \\
                   & B & 62.5 & 69.4 & 4.7  & 56.2 & 92.8 \\
Qwen2.5-VL-7B      & A & 64.0 & 78.7 & 9.0  & 53.2 & 85.3 \\
                   & B & 66.5 & 86.3 & 3.0  & 55.7 & 93.4 \\
LLaVA-OneVision-7B & A & 70.5 & 78.4 & 18.4 & 54.8 & 59.3 \\
                   & B & 70.4 & 81.3 & 7.1  & 59.6 & 87.3 \\
InternVL3-8B       & A & 73.0 & 89.9 & 0.7  & 55.5 & 99.4 \\
                   & B & 71.8 & 89.4 & 0.7  & 56.0 & 98.9 \\
\bottomrule
\end{tabular}
\end{table}

\section{Control Arms and the VSR Benchmark}
\label{app:ctrl}

\textbf{Arms.} The two control arms use CCFT v2 (the corrected loss) with unchanged
hyper-parameters, prompt and seeds, and differ only in the training and validation questions,
which are filtered by type: the non-spatial arm keeps T2--T4 (1,314 training questions with an
existing image) and the spatial-only arm keeps T1 (682), against 1,996 for the full set; the
validation file used for checkpoint selection is filtered in the same way. Both models have seeds
42, 1 and 2 for both arms. Evaluation is as in Appendix~\ref{app:ext}.

\begin{table}[h]
\centering
\small
\caption{CV-Bench accuracy (\%) with the image in the original option order, by arm (mean $\pm$ sd over seeds), exact McNemar
$p$ of the spatial-relation accuracy against the plain model for each seed, and blind
spatial-relation accuracy. LLaVA-NeXT CCFT v1 is a single run (no spread).}
\label{tab:ctrl}
\setlength{\tabcolsep}{3pt}
\begin{tabular}{llccccc}
\toprule
Model & Arm & Overall & Count & Relation & $p$ (Relation, per seed) & Blind rel. \\
\midrule
LLaVA-1.5-7B  & Plain          & 56.2 & 53.3 & 60.0 & -- & 49.7 \\
              & CCFT v1        & 47.0$\pm$2.4 & 39.1$\pm$2.7 & 57.3$\pm$3.3 & .71 / .89 / .03 & 53.5 \\
              & CCFT v2        & 56.2$\pm$0.9 & 48.2$\pm$1.0 & 66.7$\pm$0.8 & $<$.0001 (all) & 49.7 \\
              & Non-spatial    & 55.9$\pm$0.9 & 51.4$\pm$0.8 & 61.9$\pm$1.2 & .04 / .14 / .60 & 48.9 \\
              & Spatial only   & 62.3$\pm$0.5 & 54.1$\pm$0.6 & 72.9$\pm$1.5 & $<$.0001 (all) & 53.0 \\
\midrule
LLaVA-NeXT-7B & Plain          & 63.5 & 63.1 & 64.1 & -- & 49.7 \\
              & CCFT v1 (one run) & 28.9 & 12.6 & 50.1 & $<$.0001 & 48.8 \\
              & CCFT v2        & 74.9$\pm$0.9 & 64.2$\pm$2.1 & 88.7$\pm$0.6 & $<$.0001 (all) & 51.5 \\
              & Non-spatial    & 66.2$\pm$2.0 & 63.6$\pm$1.2 & 69.5$\pm$3.6 & .52 / $<$.0001 / $<$.0001 & 50.2 \\
              & Spatial only   & 76.7$\pm$0.9 & 67.3$\pm$1.3 & 89.0$\pm$0.5 & $<$.0001 (all) & 52.3 \\
\bottomrule
\end{tabular}
\end{table}

\begin{table}[h]
\centering
\small
\caption{VSR accuracy (\%, original option order, $n{=}1{,}218$; always answering ``true'' scores 51.4\%) with and without the
image, and the share of ``A'' (true) answers with the image (range over seeds). v1 on
LLaVA-NeXT is a single run.}
\label{tab:vsr}
\setlength{\tabcolsep}{4pt}
\begin{tabular}{llccc}
\toprule
Model & Arm & Image & Blind & ``A'' answers \\
\midrule
LLaVA-1.5-7B  & Plain        & 60.2 & 49.8 & 74 \\
              & CCFT v1      & 51.4$\pm$0.2 & 51.6 & 98\nobreakdash--100 \\
              & CCFT v2      & 54.5$\pm$0.4 & 51.0 & 92\nobreakdash--93 \\
              & Non-spatial  & 54.3$\pm$0.5 & 52.0 & 92\nobreakdash--94 \\
              & Spatial only & 60.7$\pm$0.7 & 49.1 & 75\nobreakdash--80 \\
\midrule
LLaVA-NeXT-7B & Plain        & 65.0 & 48.6 & 79 \\
              & CCFT v1      & 51.4 & 51.4 & 100 \\
              & CCFT v2      & 58.3$\pm$2.5 & 48.6 & 84\nobreakdash--92 \\
              & Non-spatial  & 58.4$\pm$2.5 & 48.6 & 85\nobreakdash--92 \\
              & Spatial only & 65.5$\pm$0.8 & 48.5 & 72\nobreakdash--76 \\
\bottomrule
\end{tabular}
\end{table}

\begin{table}[h]
\centering
\small
\caption{CV-Bench spatial-relation accuracy (\%) with the image in the original option order, by image source (ADE20K: 291
questions; COCO: 264 questions after excluding images shared with CompHall).}
\label{tab:src}
\setlength{\tabcolsep}{4pt}
\begin{tabular}{llcc}
\toprule
Model & Arm & ADE20K & COCO \\
\midrule
LLaVA-1.5-7B  & Plain        & 61.9 & 58.0 \\
              & CCFT v1      & 57.2 & 57.4 \\
              & CCFT v2      & 64.9 & 68.6 \\
              & Non-spatial  & 61.2 & 62.8 \\
              & Spatial only & 71.4 & 74.5 \\
\midrule
LLaVA-NeXT-7B & Plain        & 61.2 & 67.4 \\
              & CCFT v1      & 50.2 & 50.0 \\
              & CCFT v2      & 85.5 & 92.3 \\
              & Non-spatial  & 67.8 & 71.3 \\
              & Spatial only & 86.6 & 91.7 \\
\bottomrule
\end{tabular}
\end{table}

\section{Option-Order Control}
\label{app:rev}

Every two-option question (CV-Bench spatial relations after excluding images shared with
CompHall: 555 questions; VSR: 1,218) is asked in the original option order and with the two
options reversed, with the same prompts and the image. ``Averaged'' is the mean of the two
accuracies; ``consistent'' is the share of questions answered with the same option content in both
orders. The last column gives the gain of the averaged accuracy over the plain model for each seed,
with the $p$-value of a paired bootstrap over questions (5,000 resamples).

\begin{table}[h]
\centering
\small
\caption{CV-Bench spatial relations: accuracy (\%) by option order (mean over seeds), averaged
accuracy ($\pm$ sd over seeds), consistency, and per-seed gain over the plain model.}
\label{tab:rev_rel}
\footnotesize
\setlength{\tabcolsep}{2.4pt}
\begin{tabular}{llccccl}
\toprule
Model & Arm & Original & Reversed & Averaged & Consistent & Gain (per seed), $p$ \\
\midrule
LLaVA-1.5-7B  & Plain        & 60.0 & 55.1 & 57.6 & 75.7 & -- \\
              & CCFT v1      & 57.3 & 49.7 & 53.5$\pm$1.6 & 51.9 & $-$2.9 to $-$5.9; $p\le.049$ \\
              & CCFT v2      & 66.7 & 54.9 & 60.8$\pm$0.1 & 74.4 & $+$3.2 to $+$3.3; $p<.0001$ \\
              & Non-spatial  & 61.9 & 53.3 & 57.6$\pm$0.9 & 76.7 & $-$0.9 to $+$0.9; $p\ge.16$ \\
              & Spatial only & 72.9 & 56.2 & 64.5$\pm$1.3 & 68.8 & $+$5.9 to $+$8.5; $p<.0001$ \\
\midrule
LLaVA-NeXT-7B & Plain        & 64.1 & 61.4 & 62.8 & 73.2 & -- \\
              & CCFT v1 (one run) & 50.1 & 55.5 & 52.8 & 85.9 & $-$10.0; $p<.0001$ \\
              & CCFT v2      & 88.7 & 83.1 & 85.9$\pm$1.9 & 88.6 & $+$21.2 to $+$24.9; $p<.0001$ \\
              & Non-spatial  & 69.5 & 62.7 & 66.1$\pm$2.6 & 71.8 & $+$0.6 to $+$5.9; $p=.66,.002,<.0001$ \\
              & Spatial only & 89.0 & 85.1 & 87.1$\pm$0.6 & 90.9 & $+$23.6 to $+$24.8; $p<.0001$ \\
\midrule
Idefics2-8B   & Plain        & 68.5 & 59.1 & 63.8 & 65.8 & -- \\
              & CCFT v2      & 69.3 & 70.4 & 69.8$\pm$1.8 & 64.8 & $+$4.1 to $+$7.7; $p\le.002$ \\
              & Non-spatial  & 66.5 & 65.7 & 66.1$\pm$1.3 & 70.7 & $+$1.5 to $+$3.9; $p=.20,.0016,.18$ \\
              & Spatial only & 70.6 & 66.9 & 68.7$\pm$1.9 & 74.0 & $+$3.9 to $+$7.1; $p\le.002$ \\
\bottomrule
\end{tabular}
\end{table}

\begin{table}[h]
\centering
\small
\caption{VSR: accuracy (\%) by option order (mean over seeds), averaged accuracy ($\pm$ sd over
seeds), consistency, and per-seed gain over the plain model.}
\label{tab:rev_vsr}
\footnotesize
\setlength{\tabcolsep}{2.4pt}
\begin{tabular}{llccccl}
\toprule
Model & Arm & Original & Reversed & Averaged & Consistent & Gain (per seed), $p$ \\
\midrule
LLaVA-1.5-7B  & Plain        & 60.2 & 52.1 & 56.1 & 74.8 & -- \\
              & CCFT v1      & 51.4 & 51.4 & 51.4$\pm$0.0 & 98.5 & $-$4.7; $p<.0001$ \\
              & CCFT v2      & 54.5 & 51.8 & 53.1$\pm$0.2 & 93.8 & $-$2.8 to $-$3.2; $p<.0001$ \\
              & Non-spatial  & 54.3 & 51.8 & 53.0$\pm$0.2 & 93.8 & $-$2.8 to $-$3.2; $p<.0001$ \\
              & Spatial only & 60.7 & 51.9 & 56.3$\pm$0.4 & 79.4 & $-$0.1 to $+$0.6; $p\ge.12$ \\
\midrule
LLaVA-NeXT-7B & Plain        & 65.0 & 58.5 & 61.7 & 35.8 & -- \\
              & CCFT v1 (one run) & 51.4 & 52.1 & 51.8 & 99.1 & $-$10.0; $p<.0001$ \\
              & CCFT v2      & 58.3 & 65.4 & 61.9$\pm$1.3 & 78.8 & $-$1.3 to $+$1.3; $p\ge.32$ \\
              & Non-spatial  & 58.4 & 66.6 & 62.5$\pm$2.3 & 58.3 & $-$1.6 to $+$3.1; $p=.07,.002,.50$ \\
              & Spatial only & 65.5 & 64.1 & 64.8$\pm$0.3 & 61.6 & $+$2.8 to $+$3.3; $p\le.002$ \\
\midrule
Idefics2-8B   & Plain        & 76.9 & 77.3 & 77.1 & 93.3 & -- \\
              & CCFT v2      & 74.9 & 77.4 & 76.1$\pm$1.1 & 87.0 & $-$2.2 to $-$0.1; $p=.48,.0012,.88$ \\
              & Non-spatial  & 74.5 & 76.6 & 75.6$\pm$0.4 & 88.3 & $-$1.9 to $-$1.1; $p\le.044$ \\
              & Spatial only & 77.2 & 76.7 & 77.0$\pm$0.2 & 92.6 & $-$0.4 to $+$0.0; $p\ge.60$ \\
\bottomrule
\end{tabular}
\end{table}

\section{Held-Out Relation Words}
\label{app:vocab}

The horizontal-only arm is trained on the CompHall spatial training questions whose correct
option contains ``left'' or ``right'' (511 in the training file) and the vertical-only arm on those
whose correct option is above, below, over, under, top or beneath (159; 158 remained after the
training scripts skipped one question whose image file is unavailable); 17 questions with other
correct answers (for example ``behind'') are in neither arm. Validation questions used for checkpoint
selection are filtered in the same way, and all other settings are those of the full spatial-only arm.
Table~\ref{tab:vocab_cv} reports CV-Bench spatial relations split by the axis of the correct
answer, averaged over both option orders; the $p$-values are paired bootstraps over questions against
the plain model, per seed. Table~\ref{tab:vocab_comp} reports CompHall's spatial test questions in the
original option order (exact McNemar tests). An asterisk marks the held-out axis of an arm.

\begin{table}[h]
\centering
\small
\caption{CV-Bench spatial relations, accuracy (\%) averaged over both option orders, by the axis of the
correct answer (331 horizontal, 224 vertical questions); mean $\pm$ sd over three seeds and per-seed $p$.}
\label{tab:vocab_cv}
\setlength{\tabcolsep}{3pt}
\begin{tabular}{llll}
\toprule
Model & Arm & Horizontal-correct & Vertical-correct \\
\midrule
LLaVA-1.5-7B  & Plain           & 51.5 & 66.5 \\
              & All types       & 54.2$\pm$0.5; $p<.01$ (all) & 70.5$\pm$0.5; $p<.01,<.01,.07$ \\
              & Spatial only    & 58.9$\pm$1.2; $p<.01$ (all) & 72.8$\pm$1.9; $p<.01,<.01,.02$ \\
              & Horizontal only & 57.3$\pm$3.6; $p<.01$ (all) & 69.1$\pm$1.8$^*$; $p<.01,.12,.74$ \\
              & Vertical only   & 55.4$\pm$0.7$^*$; $p<.01,<.01,.01$ & 68.4$\pm$0.3; $p=.28,.37,.44$ \\
\midrule
LLaVA-NeXT-7B & Plain           & 61.0 & 65.4 \\
              & All types       & 88.4$\pm$0.4; $p<.01$ (all) & 82.1$\pm$4.4; $p<.01$ (all) \\
              & Spatial only    & 87.9$\pm$0.8; $p<.01$ (all) & 85.8$\pm$0.6; $p<.01$ (all) \\
              & Horizontal only & 85.9$\pm$0.5; $p<.01$ (all) & 78.8$\pm$0.8$^*$; $p<.01$ (all) \\
              & Vertical only   & 78.2$\pm$2.7$^*$; $p<.01$ (all) & 79.4$\pm$2.6; $p<.01$ (all) \\
\bottomrule
\end{tabular}
\end{table}

\begin{table}[h]
\centering
\small
\caption{CompHall spatial test questions, accuracy (\%) with the image in the original option order, by
the axis of the correct answer (103 horizontal, 41 vertical questions); mean $\pm$ sd over three seeds and
per-seed $p$.}
\label{tab:vocab_comp}
\setlength{\tabcolsep}{3pt}
\begin{tabular}{llll}
\toprule
Model & Arm & Horizontal-correct & Vertical-correct \\
\midrule
LLaVA-1.5-7B  & Plain           & 41.7 & 39.0 \\
              & All types       & 52.4$\pm$1.0; $p<.01,.02,<.01$ & 35.0$\pm$3.7; $p=.62,1.0,.38$ \\
              & Spatial only    & 55.7$\pm$1.1; $p<.01,.04,.01$ & 45.5$\pm$3.7; $p=1.0,.51,.22$ \\
              & Horizontal only & 50.8$\pm$2.0; $p<.01,.09,.19$ & 40.7$\pm$3.7$^*$; $p=.62,1.0,1.0$ \\
              & Vertical only   & 38.2$\pm$1.1$^*$; $p=.12,.61,.55$ & 45.5$\pm$1.4; $p=.69,.51,.51$ \\
\midrule
LLaVA-NeXT-7B & Plain           & 42.7 & 46.3 \\
              & All types       & 63.1$\pm$1.9; $p<.01$ (all) & 49.6$\pm$5.1; $p=.80,1.0,.58$ \\
              & Spatial only    & 61.8$\pm$1.1; $p<.01$ (all) & 56.1$\pm$2.4; $p=.61,.30,.45$ \\
              & Horizontal only & 63.4$\pm$1.5; $p<.01$ (all) & 41.5$\pm$4.9$^*$; $p=.42,.80,1.0$ \\
              & Vertical only   & 36.6$\pm$2.2$^*$; $p=.30,.08,.33$ & 68.3$\pm$2.4; $p=.02,.06,.01$ \\
\bottomrule
\end{tabular}
\end{table}

\section{Extended Background}
\label{app:background}

\subsection*{MLLM Architecture and Visual Encoding}

\begin{figure}[htb]
\centering
\includegraphics[width=0.82\linewidth]{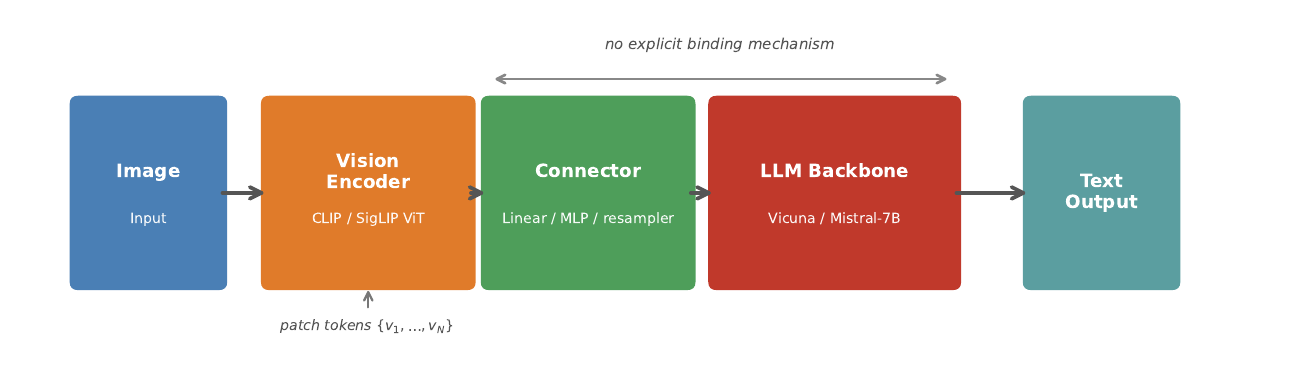}
\caption{Standard MLLM pipeline. Patch tokens from the vision encoder are
projected into the LLM's embedding space, and no component is designed for binding;
object-feature attribution is handled by LLM self-attention layers.}
\label{fig:arch_pipeline}
\end{figure}

Modern MLLMs follow a three-stage pipeline
(Figure~\ref{fig:arch_pipeline}): a \textbf{vision encoder} maps an input image
to a sequence of visual tokens; a \textbf{connector module} projects or
cross-attends those tokens into the language model's embedding space; and an
\textbf{LLM backbone} generates text conditioned on the fused multimodal context.

\textbf{Vision encoders.} The standard choice is a CLIP ViT~\cite{radford2021clip}
trained on image--text contrastive objectives. A ViT divides the image into a
$P{\times}P$ grid of patches, processes each through self-attention layers, and
outputs a sequence of patch tokens $\{v_1, \ldots, v_N\}$. Patch tokens carry
local appearance information; spatial relationships between patches are encoded
only through learned position embeddings, which do not explicitly represent
object-level spatial relations.

\textbf{Connector modules.} LLaVA~\cite{liu2023llava} projects each patch token with a linear layer
(LLaVA-1.5~\cite{liu2024improved} uses a two-layer MLP). InstructBLIP and BLIP-2~\cite{dai2023instructblip,li2023blip2}
use a Q-Former: $K$ learned query vectors that cross-attend over all $N$ patch
tokens, producing a fixed-length visual sequence. Idefics2~\cite{laurenccon2024idefics2}
uses a perceiver resampler. LLaVA-NeXT~\cite{liu2024llavanext} tiles the image
at multiple resolutions before encoding, increasing the number of patch tokens
per image.

\textbf{The binding bottleneck.} In all of these architectures, the attribution
of spatial and attribute information to specific objects is handled by LLM
self-attention layers. These layers were trained on text and have no structural
inductive bias toward enforcing feature-to-object consistency. There is no
explicit operation that associates a colour token with its object rather than a
neighbouring one.
This bottleneck becomes more acute as scene complexity increases.
For a single-object scene, feature attribution is unambiguous.
For a scene with $K$ objects of varying colours, sizes, and positions, the
model must resolve $K$-way attribution across every feature dimension using
only the implicit structure of LLM self-attention weights---a task those
weights were not pretrained to perform explicitly.
The expected difficulty therefore scales with object count and attribute
diversity; Section~\ref{sec:clevr} measures how existence-question errors change
with object count, although those questions do not themselves require binding.

\subsection*{The Binding Problem in Perception}

The \emph{binding problem}~\cite{treisman1980feature,treisman1996binding} refers
to the question of how the brain combines independently processed features
(colour, shape, position) into correctly attributed object representations.
Treisman's Feature Integration Theory proposes that a serial focal attention
operation selects one location at a time and assembles all features at that
location into a coherent percept. Without this step, features are
susceptible to \emph{illusory conjunctions}: incorrect attributions in which a
feature from one object is assigned to another (\eg, perceiving a red circle and
a blue square as a blue circle and a red square).

MLLMs process all patch tokens in parallel without an explicit location-selection
step, and may produce the same class of attribution errors: correctly
representing that a scene contains ``red,'' ``sphere,'' and ``left,'' but
binding them incorrectly.

\subsection*{A Taxonomy of MLLM Hallucination}

We organise MLLM hallucination into four types based on which binding step fails:

\begin{itemize}
\item \textbf{Object-existence (T0).} The model reports an object not present
      in the image. This is the target of POPE and most prior benchmarks.
      It does not require a compositional binding error; a model can fail here
      even with correct attribute-to-object binding.

\item \textbf{Attribute-binding (T2).} The model correctly identifies objects
      but assigns an attribute (colour, size, texture) to the wrong one.
      \eg, ``a large red sphere'' when the sphere is blue and the cube is red.

\item \textbf{Spatial-relation (T1).} The model mis-reports the relative or
      absolute position of an object.

\item \textbf{Part--whole and role binding (T4).} The model assigns the wrong
      roles in a part--whole or possession relation, \eg, ``the pants are wearing
      the man'' for a man wearing pants.

\item \textbf{Counting (T3).} The model reports an incorrect count of a target
      category. Correct counting requires identifying every instance of the
      category, so the number of required bindings scales with scene population
      rather than being fixed.
\end{itemize}

CompHall targets T1, T2, T3 (counting, which requires per-instance
identification), and T4. T0 is excluded by design: every CompHall question
involves only objects that are present in the image. This separation matters
because a model can do well on a T0 benchmark (POPE accuracy 0.82\nobreakdash--0.87 for
LLaVA-1.5-7B) while performing near chance on T1 and T3, as our results show.

\section{Accuracy Heatmap}
\label{app:heatmap}

\begin{figure}[h]
\centering
\includegraphics[width=0.82\linewidth]{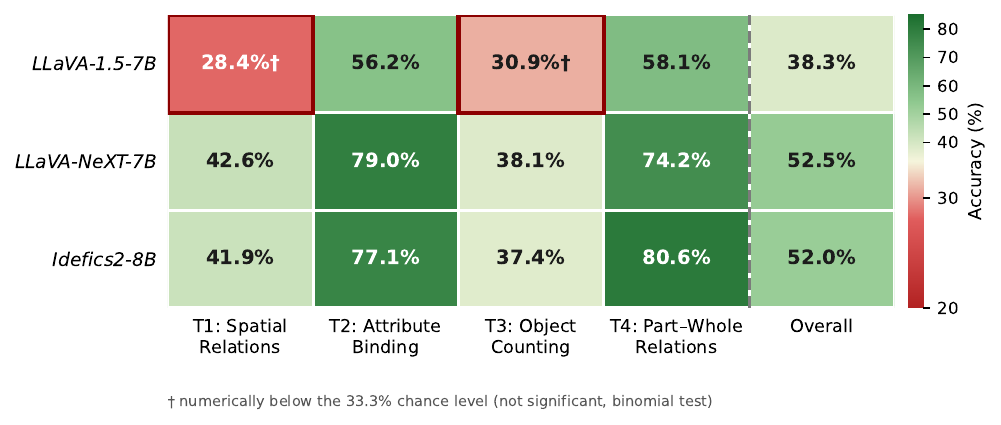}
\caption{CompHall accuracy by model and question type (first prompt wording; the three recent
models are in Table~\ref{tab:recent}). Cells numerically below
the 33.3\% chance level are marked ($\dagger$); none is significantly below chance
(two-sided binomial test). LLaVA-NeXT-7B and Idefics2-8B score 42.6\% and 41.9\%
on T1 but 74\nobreakdash--81\% on T2 and T4.}
\label{fig:heatmap}
\end{figure}

\section{Idefics2-8B}
\label{app:idefics}

Idefics2-8B (SigLIP vision encoder, perceiver resampler, Mistral-7B language model) is fine-tuned with the corrected
loss of Section~\ref{sec:method} (answer-token score; contrastive and ranking terms with the weights of
Table~\ref{tab:hyperparams}), with LoRA of rank 16 ($\alpha{=}32$, dropout 0.05) on the q, k, v and o projections of the
language model only (13.6M trainable parameters), learning rate $2\times10^{-5}$ (AdamW, weight decay 0.01, cosine
schedule with 10\% warm-up), 3 epochs, gradient clipping at 1.0, fp16 autocast with fp32 LoRA weights, one question per
optimizer step (its correct answer and the two negatives) and no image splitting (64 visual tokens per image); the last
checkpoint is used. The three arms are trained on all question types (1,996 questions), on the non-spatial types
(1,314) and on spatial questions (682), with seeds 42, 1 and 2. Evaluation uses the letter prompt of the other models
(Section~\ref{sec:baseline}); the plain model scores 52.2\% with the image under this implementation (54.6\% with
the parser of Section~\ref{sec:baseline}). Table~\ref{tab:idef_comp} gives CompHall by type and
Table~\ref{tab:idef_ext} CV-Bench and VSR in the original option order; the order-averaged results are in
Tables~\ref{tab:rev_rel} and~\ref{tab:rev_vsr}.

\begin{table}[h]
\centering
\footnotesize
\caption{Idefics2-8B on CompHall: accuracy (\%) with the image and with a gray image (mean $\pm$ sd over three seeds).
With the image, no arm differs from the plain model overall (McNemar $p\ge0.40$ in every seed) or on T1 ($p\ge0.19$).}
\label{tab:idef_comp}
\setlength{\tabcolsep}{4pt}
\begin{tabular}{llccccc}
\toprule
Input & Arm & T1 & T2 & T3 & T4 & Overall \\
\midrule
With image & Plain        & 42.6 & 73.3 & 38.8 & 87.1 & 52.2 \\
           & All types    & 39.4$\pm$2.2 & 79.0$\pm$1.9 & 39.8$\pm$1.1 & 83.9$\pm$3.2 & 52.6$\pm$0.4 \\
           & Non-spatial  & 40.1$\pm$1.4 & 79.4$\pm$2.9 & 41.5$\pm$1.1 & 84.9$\pm$1.9 & 53.6$\pm$0.8 \\
           & Spatial only & 42.1$\pm$1.0 & 75.2$\pm$1.6 & 35.3$\pm$1.9 & 84.9$\pm$6.7 & 51.2$\pm$0.5 \\
\midrule
Blind      & Plain        & 46.6 & 52.4 & 30.9 & 87.1 & 45.9 \\
           & All types    & 41.2$\pm$1.4 & 59.7$\pm$3.8 & 39.3$\pm$1.1 & 90.3$\pm$0.0 & 48.8$\pm$1.0 \\
           & Non-spatial  & 43.0$\pm$2.6 & 61.9$\pm$1.9 & 38.6$\pm$0.8 & 92.5$\pm$1.9 & 49.9$\pm$0.6 \\
           & Spatial only & 41.2$\pm$2.9 & 53.3$\pm$0.0 & 30.7$\pm$1.7 & 87.1$\pm$0.0 & 44.1$\pm$1.6 \\
\bottomrule
\end{tabular}
\end{table}

\begin{table}[h]
\centering
\footnotesize
\caption{Idefics2-8B on CV-Bench (relation and counting) and VSR in the original option order: accuracy (\%) with the
image (mean $\pm$ sd over three seeds; per-seed exact McNemar $p$ against the plain model in brackets) and with a gray image
(means).}
\label{tab:idef_ext}
\setlength{\tabcolsep}{2.4pt}
\begin{tabular}{lcccccc}
\toprule
 & \multicolumn{3}{c}{With image} & \multicolumn{3}{c}{Blind} \\
\cmidrule(r){2-4}\cmidrule(r){5-7}
Arm & Relation & Counting & VSR & Relation & Counting & VSR \\
\midrule
Plain        & 68.5 & 61.5 & 76.9 & 55.9 & 7.7 & 48.6 \\
All types    & 69.3$\pm$2.4 (.17, .68, .55) & 59.0$\pm$0.8 (.048, .29, .10) & 74.9$\pm$1.3 (.044, .0001, .30) & 55.6 & 28.3 & 49.7 \\
Non-spatial  & 66.5$\pm$1.8 (.16, 1.0, .23) & 58.6$\pm$0.7 (.016, .12, .14) & 74.5$\pm$0.5 (.005, .009, .0003) & 57.3 & 27.3 & 49.2 \\
Spatial only & 70.6$\pm$1.4 (.35, .08, .78) & 65.4$\pm$0.4 (.001, .0002, .0009) & 77.2$\pm$0.5 (.90, .38, .92) & 53.8 & 5.8 & 48.7 \\
\bottomrule
\end{tabular}
\end{table}

\section{POPE Figure}
\label{app:popefig}

\begin{figure}[h]
\centering
\includegraphics[width=0.82\linewidth]{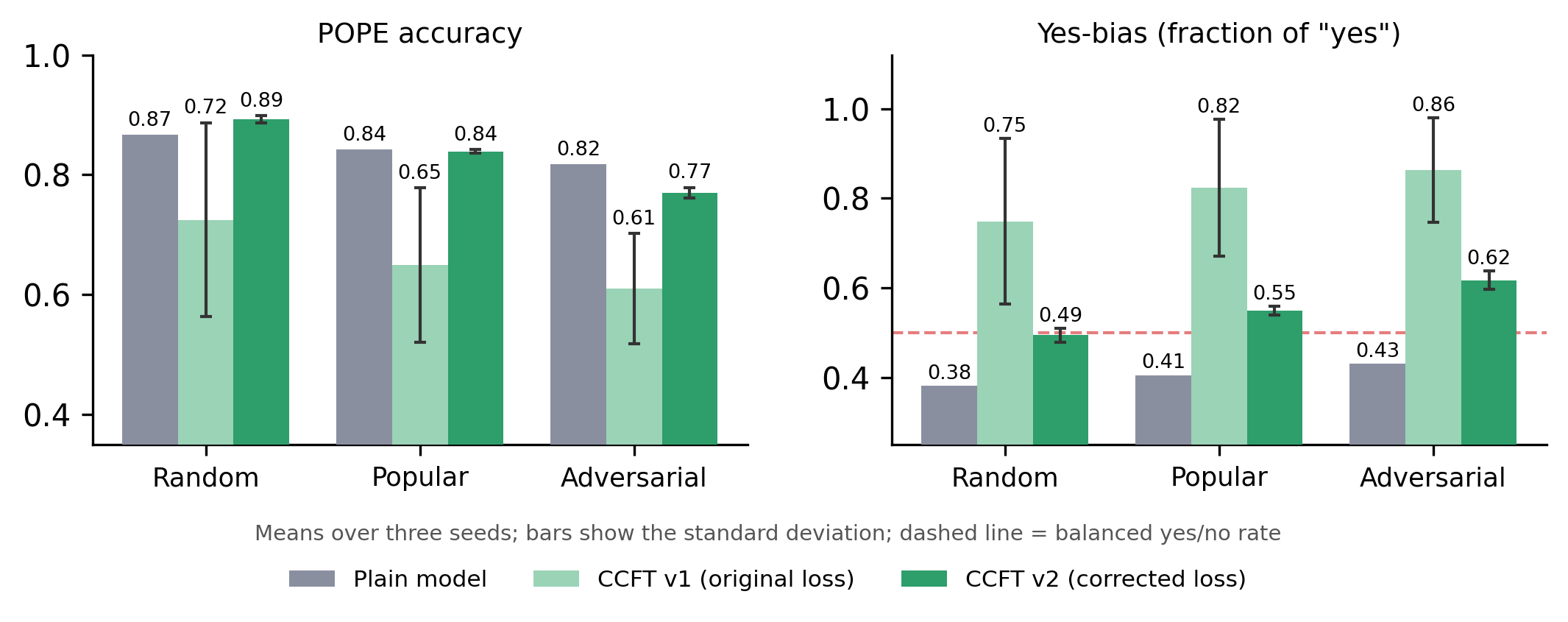}
\caption{POPE accuracy and yes-bias for the plain model, CCFT v1 and CCFT v2
(mean over three seeds; bars show the standard deviation). The dashed line marks a
balanced yes/no rate.}
\label{fig:pope}
\end{figure}

\end{document}